\documentclass[11pt]{article}
\usepackage[utf8]{inputenc}
\usepackage{verbatim}
\usepackage{pdfpages}
\usepackage{amsmath, array}
\usepackage{physics}
\usepackage{natbib}
\usepackage{booktabs}       
\usepackage{multirow}
\usepackage{amsfonts}      
\usepackage{nicefrac}      
\usepackage{microtype}     
\usepackage{parskip}
\usepackage{overpic}
\usepackage{subcaption}
\usepackage{authblk}
\usepackage{array}
\usepackage[colorlinks]{hyperref}
\usepackage{tikz}
\usetikzlibrary{arrows.meta,positioning,fit,calc,backgrounds}

\hypersetup{
	linkcolor = green, 
	citecolor = magenta 
}

\usepackage[margin=1in]{geometry}
\usepackage{natbib}
\usepackage{verbatim}
\usepackage{pdfpages}
\usepackage{amsmath, array}
\usepackage{booktabs}       
\usepackage{multirow}
\usepackage{amsfonts}       
\usepackage{nicefrac}       
\usepackage{microtype}      
\usepackage{graphicx}
\usepackage{overpic}

\usepackage[table]{xcolor}
\definecolor{iFLEXrow}{RGB}{255,239,219}
\usepackage{tikz}
\usepackage{pict2e}
\usepackage{subcaption}

\usetikzlibrary{arrows.meta,positioning,fit,calc,backgrounds}

\newcommand{\model}{\texttt{iFLEX}}

\usepackage{flafter}
\usepackage{placeins}

\title{\LARGE Generative Refinement Learning for Temporal Interpolation in Scientific Machine Learning}

\author[1]{Pu Ren\thanks{Email: \texttt{ren.pu@northeastern.edu}}}
\author[1,2,3]{Michael W. Mahoney}
\author[1,2]{N. Benjamin Erichson}

\affil[1]{Lawrence Berkeley National Lab}
\affil[2]{International Computer Science Institute}
\affil[3]{University of California, Berkeley}

\begin{document}

\maketitle

\begin{abstract}
Computationally expensive simulations and real-world observations often provide data at a lower temporal resolution than downstream tasks require. Averaging observations at fixed grid points can introduce errors, and motion estimates alone do not determine how the physical field values change. In this work, we study whether learning to correct a simple numerical interpolation improves the reconstruction of intermediate scientific fields. Our model, \textbf{i}nterpolative \textbf{FL}ow \textbf{EX}pert (\model{}), first estimates the missing field by taking a weighted average of the two observations at each grid point, giving more weight to the observation closer to the requested time. During training, subtracting this estimate from an observed intermediate field gives the residual that a conditional diffusion model learns to generate. Both observations condition the model, together with the target time, the interval between observations, and available physical parameters. To reconstruct an unobserved intermediate field, the generated residual is added to the weighted average. We evaluate \model{} on Navier--Stokes Kraichnan turbulence, Shanghai radar precipitation, and sea surface temperature. Across these datasets, \model{} achieves lower reconstruction error than the evaluated baselines, including optical-flow and diffusion models developed for video frame interpolation in computer vision. In a controlled turbulence comparison at a Reynolds number of 36,000, learning the residual reduces relative Frobenius norm error from about 13.7\% for direct field generation to 10.5\%, with the diffusion backbone and training setup held fixed. Code is available at \url{https://github.com/paulpuren/iflex}.
\end{abstract}

\begin{figure}[!t]
    \centering
    \begin{tikzpicture}
        \node[anchor=south west, inner sep=0] (img) at (0,0)
        {\includegraphics[width=0.85\textwidth]{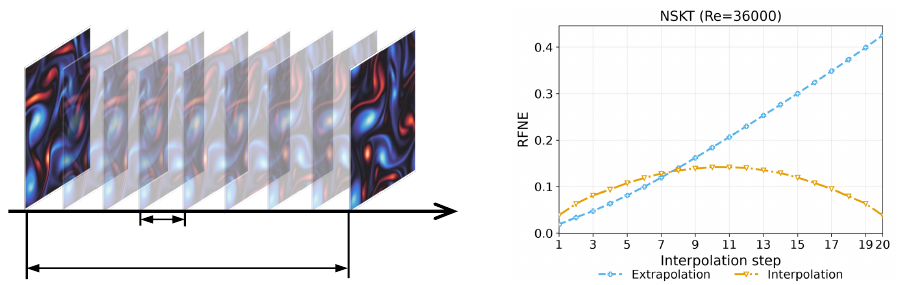}};

        \begin{scope}[x={(img.south east)}, y={(img.north west)}]

            \node at (0.001, 0.95) {\textbf{a.}};
            \node at (0.55, 0.95) {\textbf{b.}};
            
            \node at (0.21, 0.11) {\small $\Delta t$};
            \node at (0.18, 0.185) {\small $\delta t$};
          
        \end{scope}
    \end{tikzpicture}
    \caption{Interpolation using two observed states achieves lower reconstruction error than the tested one-endpoint predictor on NSKT.
    (a) Observations are separated by a time interval $\Delta t$, and intermediate fields are reconstructed at a finer temporal spacing $\delta t$. Unlike prediction from an initial state, interpolation uses observations from both ends of the interval. (b) At $Re=36000$, iFLEX achieves lower relative Frobenius norm error (RFNE) than the one-endpoint predictor (Model A) across the tested intermediate times. Over the full evaluation set, the RFNE is $10.5\%$ for iFLEX and $20.4\%$ for Model A. Reconstruction errors are generally larger near the midpoint, where the requested state is farthest in time from both observations.}
    \label{fig:interp_vs_extrap}
\end{figure}

\section{Introduction}\label{s:intro}

High-fidelity fluid simulations often require very small time steps to accurately resolve the underlying dynamics. However, storing the complete flow field after every time step can be prohibitively expensive, particularly for large three-dimensional simulations. To reduce storage and data transfer costs, simulations therefore save only selected snapshots, leaving gaps in the temporal record. A similar problem arises with real-world observations, where the temporal resolution is limited by measurement frequency. For example, meteorological products such as ERA5 provide hourly weather data~\citep{hersbach2020era5}, whereas applications such as flash flood modeling may require precipitation information at intervals of only a few minutes. In both cases, subsequent analyses may need physical fields at times for which no data were stored or observed. Temporal interpolation addresses this problem by estimating the missing states from two observations that bracket the requested time. Unlike forecasting, which must predict the future evolution from earlier observations, interpolation can also use the later observation to constrain the reconstruction. Figure~\ref{fig:interp_vs_extrap} illustrates this distinction. Accurate temporal interpolation could therefore allow simulations to save fewer snapshots while retaining access to estimates of intermediate states, and help make observational records available at finer temporal intervals.

Reconstructing these intermediate states is challenging because physical fields can change considerably between observations. The simplest approach, linear interpolation, computes a weighted average of the two observed fields at each grid point. While inexpensive, this can produce inaccurate results when structures move or deform. For example, if a vortex changes position between two observations, linear interpolation may produce two weakened structures near the observed locations rather than a vortex at its intermediate position. A related problem has been studied extensively in video frame interpolation (VFI)~\citep{dong2023video}, where optical-flow methods estimate the motion of image content between frames~\citep{horn1981determining,baker2011database}. Many of these methods approximate motion using linear~\citep{jiang2018super} or quadratic~\citep{xu2019quadratic,liu2020enhanced} trajectories. However, scientific fields can undergo changes beyond the displacement of individual structures. A vortex may stretch and diffuse as it moves, while a precipitation cell may intensify or dissipate. Reconstructing the intermediate field therefore requires estimating how its physical values evolve, in addition to where structures move. More recent approaches use neural networks to predict intermediate frames directly~\citep{kye2025acevfi}, or generative models that operate in pixel space~\citep{jain2024video} or learned latent representations~\citep{danier2024ldmvfi}. Although these methods provide useful alternatives, their performance on natural video does not necessarily translate to scientific fields governed by different dynamics. Thus, we study how temporal interpolation should be formulated for scientific data, and whether learning to correct a simple numerical interpolation can improve reconstruction accuracy.

\begin{figure}[!t]
    \centering
    \begin{tikzpicture}
        \node[anchor=south west, inner sep=0] (img) at (0,0)
        {\includegraphics[width=1.0\textwidth]{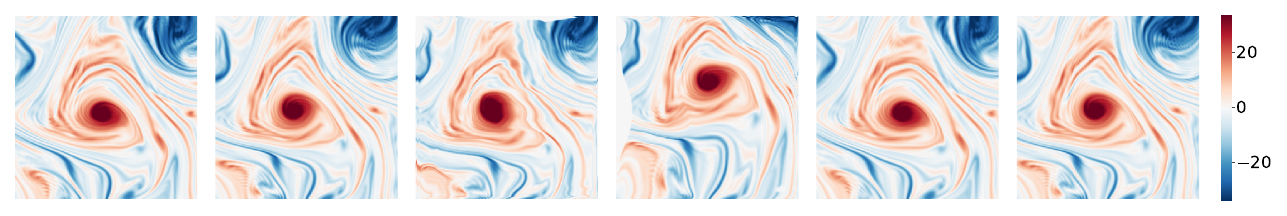}};

        \begin{scope}[x={(img.south east)}, y={(img.north west)}]
            \node at (0.16, 1.05) {Ground truth};
            \node at (0.085, -0.01) {\small$t=0.005~(s)$};
            \node at (0.24, -0.01) {\small$t=0.02~(s)$};
            \draw[gray, dashed, line width=1pt] (0.317, -0.1) -- (0.317, 1.1);

            \node at (0.47, 1.05) {Super SloMo};
            \node at (0.395, -0.01) {\small$t=0.005~(s)$};
            \node at (0.55, -0.01 ) {\small$t=0.02~(s)$};
            \draw[gray, dashed, line width=1pt] (0.627, -0.1) -- (0.627, 1.1);

            \node at (0.785, 1.05) {\model};
            \node at (0.705, -0.01) {\small$t=0.005~(s)$};
            \node at (0.860, -0.01) {\small$t=0.02~(s)$};

            \draw[red, line width=1.2pt, rounded corners=1pt]
                (0.063, 0.38) rectangle (0.1, 0.58);

            \draw[red, line width=1.2pt, rounded corners=1pt]
                (0.213, 0.38) rectangle (0.25, 0.58);
                
            \draw[red, line width=1.2pt, rounded corners=1pt]
                (0.265, 0.66) rectangle (0.31, 0.911);

            \draw[red, line width=1.2pt, rounded corners=1pt]
                (0.363, 0.38) rectangle (0.4, 0.58);
            
            \draw[red, line width=1.2pt, rounded corners=1pt]
                (0.513, 0.38) rectangle (0.55, 0.58);
            
            \draw[red, line width=1.2pt, rounded corners=1pt]
                (0.575, 0.66) rectangle (0.62, 0.911);

            \draw[red, line width=1.2pt, rounded corners=1pt]
                (0.683, 0.38) rectangle (0.72, 0.58);

            \draw[red, line width=1.2pt, rounded corners=1pt]
                (0.838, 0.38) rectangle (0.875, 0.58);
            
            \draw[red, line width=1.2pt, rounded corners=1pt]
                (0.885, 0.66) rectangle (0.93, 0.911);

        \end{scope}
    \end{tikzpicture}

    \caption{
    In these examples, \model{} more closely reproduces the vortex structures in the ground-truth NSKT fields than Super SloMo. The panels compare the ground truth with reconstructions from Super SloMo and \model{} at $t=0.005\,\mathrm{s}$ and $t=0.02\,\mathrm{s}$, using the same two endpoint observations. Red boxes highlight differences in vortex shape and fine-scale structures. Each intermediate field is reconstructed independently.
    }
    \label{fig:teaser_fig}
\end{figure}

In this work, we introduce \model{} (interpolative Flow Expert), a conditional diffusion model for reconstructing physical fields between two observations. Our approach combines a simple numerical interpolation with a learned correction. For a requested intermediate time, we first compute a linear estimate by taking a weighted
average of the two observed fields at each grid point. Although this estimate is inexpensive to compute, it can misrepresent the position, shape, and magnitude of evolving structures. We therefore train a conditional diffusion model~\citep{ho2020denoising,song2021ddim} to learn the difference between the linear estimate and the observed intermediate field, which we call the residual. We build on the conditioning approach of FLEX~\citep{erichson2025flex},
adapting it to use both observations, the requested time,
the interval between observations, and available physical parameters. During reconstruction, the model generates the residual from these inputs and adds it to the linear estimate to obtain the reconstructed field. For example, if a vortex moves between observations, the correction can remove values left at its observed positions and restore the vortex at its intermediate location, while also accounting for changes in its shape and vorticity. This allows the model to reconstruct changes that simple linear interpolation cannot capture. The initial estimate requires no additional numerical solver or knowledge of the governing equations, making the approach applicable to both simulated and observed physical fields.

We evaluate \model{} on Navier--Stokes Kraichnan turbulence
(NSKT)~\citep{ren2025superbench}, Shanghai radar
precipitation~\citep{chen2020deep}, and sea surface
temperature (SST)~\citep{xing2023helmfluid}.
These datasets cover different types of physical evolution,
from rapidly changing turbulent structures and precipitation
patterns to comparatively smooth variations in ocean temperature. Across all three datasets, \model{} achieves lower reconstruction error than the evaluated baselines, including optical-flow and diffusion methods developed for video frame interpolation (VFI). Figure~\ref{fig:teaser_fig} illustrates the differences in reconstructing turbulent flow structures. To understand the contribution of the residual formulation, we also compare \model{} with a diffusion model that generates the intermediate field directly. At Reynolds number $Re=36000$, linear interpolation alone produces a relative Frobenius norm error (RFNE) of $43.96\%$. Direct generation reduces this error to $13.73\%$, while learning
the residual further reduces it to $10.48\%$. Since both diffusion models use the same architecture, observations, conditioning, and training procedure, this comparison isolates the effect of the learning target. The results show that even an inaccurate linear interpolation can provide a useful reference for learned reconstruction. Additional experiments examine unseen Reynolds numbers, different temporal gaps, and a held-out ocean region. Once trained, \model{} can reconstruct intermediate fields at arbitrary requested times within an observed interval, providing
estimates of missing simulation snapshots or observational
records without rerunning the underlying simulation.

Overall, our contributions are summarized below:
\begin{itemize}
    \item We introduce a temporal interpolation method that combines a simple numerical estimate with a learned generative correction. Given two observed fields, \model{} uses linear interpolation to estimate the missing state and trains a conditional diffusion model to reconstruct the remaining error. This allows the model to account for changes in the position, shape, and values of physical structures without requiring a governing-equation solver.

    \item We adapt the FLEX diffusion architecture to reconstruct intermediate states from two observations. The model combines spatial features from both observed fields with information about the requested time, the interval between observations, and available physical parameters. Training across different temporal gaps allows the model to reconstruct intermediate
    fields at different temporal resolutions.

    \item We evaluate \model{} on simulated turbulence, radar
    precipitation, and sea surface temperature, comparing it with linear interpolation, supervised learning, and optical-flow and generative methods developed for video frame interpolation. Across all three datasets, \model{} achieves lower reconstruction error than the evaluated baselines. A controlled experiment further shows that learning the residual improves reconstruction
    accuracy compared with directly generating the intermediate field. Moreover, we study unseen physical parameters, different temporal gaps, and a held-out geographic region.
\end{itemize}

\section{Related Work}\label{sec:related_work}

In this section, we review existing approaches to reconstructing missing information in scientific data, with a particular focus on temporal interpolation. We first discuss methods developed for scientific fields, followed by video frame interpolation techniques from computer vision. We then review diffusion-based approaches for modeling physical systems, including the FLEX architecture on which our method builds.

\textbf{Reconstruction of Scientific Fields.}
Recovering missing information from scientific data has been studied in several settings, depending on what information is available. When measurements are sparse in space, neural networks can reconstruct full flow fields from a limited number of sensors ~\citep{erichson2020shallow,fukami2021global}. Related methods use sparse observations to predict how physical systems evolve over time~\citep{kutz2024shallow,song2024forecasting,lyu2024wavecastnet}. Another line of research focuses on spatial super-resolution, where the goal is to recover fine-scale structures from coarse simulation data~\citep{fukami2019super,fukami2021machine}. These methods have been extended to space-time super-resolution
~\citep{esmaeilzadeh2020meshfreeflownet,ren2022physics},
while learned PDE solvers have been developed to improve
the accuracy and efficiency of simulations on coarse grids
~\citep{kochkov2021machine,sun2023neural}. Similar problems arise in climate modeling, where downscaling methods reconstruct fields at finer spatial resolutions~\citep{jiang2023efficient,bassetti2024diffesm}.
Our work focuses on temporal interpolation, where the spatial
grid is already observed at two times and the goal is to
reconstruct the missing intermediate state. FLINT~\citep{gadirov2025flint} addresses this problem by estimating flow fields for scientific datasets and using them to interpolate scalar fields. HyperFLINT~\citep{gadirov2025hyperflint} extends this approach by incorporating simulation parameters through a
hypernetwork. These methods demonstrate how learned motion
estimation can support temporal reconstruction in scientific
data. In our work, we investigate a different approach that
starts from a simple linear interpolation and learns to correct its errors directly.

\textbf{Video Frame Interpolation.}
Temporal interpolation has also been studied extensively in
computer vision, where video frame interpolation (VFI) aims
to reconstruct intermediate images from two observed frames.
Many methods estimate optical flow between the frames and
use the resulting motion fields to reconstruct image content
at the requested time. Super SloMo~\citep{jiang2018super},
for example, uses a linear motion approximation, while
quadratic interpolation methods account for more complex
motion trajectories~\citep{xu2019quadratic,liu2020enhanced}.
Although these methods can accurately reconstruct moving
objects, changes in appearance and large displacements remain
challenging~\citep{kye2025acevfi}. More recent approaches
use diffusion models to generate intermediate frames directly
in pixel space~\citep{jain2024video} or in a compressed
latent representation~\citep{danier2024ldmvfi}. Other methods incorporate motion estimates into the generation process~\citep{huang2024motion}, or use residual
diffusion to correct an initial interpolation~\citep{chavez2025time}. These approaches provide relevant alternatives for scientific temporal interpolation. However, the evolution of a physical field involves changes in the underlying quantities as well as the movement of structures. For example, the displacement of a vortex does not fully describe how its shape and vorticity change over time. We therefore evaluate methods developed for video interpolation on scientific fields and investigate whether a learned correction to numerical interpolation provides a more accurate reconstruction.

\textbf{Diffusion Models for Scientific Data.}
Diffusion models~\citep{ho2020denoising} have also been applied to reconstructing and predicting scientific fields. Dyffusion~\citep{ruhling2023dyffusion} and its spherical extension~\citep{ruhling2024probablistic} use diffusion-based methods for forecasting fluid dynamics and climate variables. Related generative approaches have been developed for weather ensembles~\citep{li2024generative}, time series~\citep{naiman2024iclr}, and earthquake ground motions
~\citep{ren2026learning}. Conditional diffusion models
have also been used to reconstruct turbulent fields from
limited measurements~\citep{du2024conditional} and to
downscale climate data~\citep{lopez2025dynamical}. More recently, flow matching has also been used for forecasting
fluid dynamics, with \citet{lim2025elucidating} showing that
the choice of probability path can substantially affect
prediction accuracy across different PDE systems.
Of particular relevance to our work, DiffCast~\citep{yu2024diffcast} combines deterministic precipitation forecasts with diffusion-based residual corrections. FLEX~\citep{erichson2025flex} similarly uses residual diffusion for modeling spatiotemporal physical systems and introduces a conditioning architecture that combines information from observed fields with the denoising network. We build on this architecture and adapt its conditioning to interpolation between two observed states. Endpoint-conditioned diffusion has also been explored in scientific imaging. For example, FB-Diff~\citep{you2025fb} uses Fourier-based motion information to reconstruct intermediate states in four-dimensional medical imaging. These studies establish residual learning and conditional diffusion as useful approaches for reconstructing evolving fields. Our contribution is to investigate a specific formulation for scientific temporal interpolation, where a weighted average of the two observations provides the initial estimate and diffusion learns the remaining correction. We evaluate this approach across physical datasets and compare it with direct generation of the intermediate field under matched model and training settings.
\section{Problem Formulation: Temporal Interpolation from Two Observed States}
\label{sec:problem_formulation}

In this section, we formalize the temporal interpolation problem and describe the information available for reconstructing intermediate physical fields. Let $\mathbf{u}(t)\in \mathbb{R}^{H\times W}$ denote a physical field defined on a spatial grid, such as the vorticity of a fluid flow, precipitation intensity, or sea surface temperature.
We assume that the field is observed at two times, $t_0$ and $t_1$, with $t_0<t_1$, and denote the corresponding observations by $\mathbf{u}_0=\mathbf{u}(t_0)$ and
$\mathbf{u}_1=\mathbf{u}(t_1)$. Our goal is to reconstruct
the unknown field $\mathbf{u}_\tau$ at an intermediate
time $t_\tau$ between these observations. 
Unlike forecasting, which predicts future states from earlier
observations, temporal interpolation can also use information
from the end of the interval to constrain the reconstruction.
However, the two observations do not necessarily reveal how
the system evolved between them. For example, observing a
vortex at two different positions does not tell us exactly
how its shape and vorticity changed during the intervening time.
The accuracy of the reconstruction therefore depends on
the temporal spacing of the observations. In this work, we
assume that the observations are sufficiently close in time
to constrain the intermediate evolution. Short-lived events
that occur entirely between observations, such as a brief
rainfall burst, may leave no evidence in either endpoint
and cannot be reconstructed from these observations alone.

The reconstruction depends on both the requested intermediate time and the time elapsed between the two observations. We denote the duration of the observation interval by $\Delta t=t_1-t_0$ and specify the requested time by its relative position $\tau\in(0,1)$ within this interval
\begin{equation}
\label{eq:setup}
    t_\tau = t_0 + \tau \Delta t.
\end{equation}
For example, $\tau=0.5$ corresponds to the midpoint between
the observations. However, reconstructing the midpoint of a
ten-minute interval is different from reconstructing the
midpoint of a one-hour interval. Although the relative position is identical, the system has more time to evolve over the longer interval, potentially leading to larger changes in the position, shape, and values of physical structures. The model therefore needs to know the actual time separation $\Delta t$, in addition to the interpolation ratio $\tau$. The evolution may also depend on physical parameters, such as the Reynolds number in fluid
simulations. When these parameters are available, we include
them as additional conditioning information
$\boldsymbol{\lambda}$. In the following, we collect the two
temporal variables in $\mathbf{p}=(\tau,\Delta t)$.

We formulate temporal interpolation as learning the relationship
between the two observed fields and the unknown intermediate state.
As in previous learning-based approaches to scientific temporal
interpolation~\citep{gadirov2025flint}, we assume that training
examples contain all three fields,
$(\mathbf{u}_0,\mathbf{u}_\tau,\mathbf{u}_1)$,
so that the intermediate state provides a known reconstruction
target. At inference time, only the two endpoint observations
and their associated conditions are available. We seek to model
the conditional distribution
\begin{equation}
\label{eq:learning_obj}
    p\left(
    \mathbf{u}_\tau
    \mid
    \mathbf{u}_0,
    \mathbf{u}_1,
    \boldsymbol{\lambda},
    \mathbf{p}
    \right),
\end{equation}
and use the learned model to obtain a reconstruction
$\widehat{\mathbf{u}}_\tau$ of the missing field.
The conditional formulation accounts for the fact that the
available observations may not contain enough information
to uniquely determine the intermediate state. For instance,
a deterministic fluid simulation is fully determined by its
complete initial state and governing equations, but the
observations supplied to the model may cover only part of
the spatial domain. Flow outside the observed region can
influence the intermediate state without being directly
represented in the endpoint fields. Similar challenges arise
when reconstructing turbulent flows from incomplete
measurements~\citep{du2024conditional}. The conditional
distribution therefore describes possible intermediate
states given the available observations, without assuming
that the underlying physical dynamics are stochastic.
To reconstruct the missing field, we could train a model
to generate the intermediate state directly. However,
the two observed fields provide a simple initial
estimate via linear interpolation. In the following
section, we develop a method that uses this estimate as
a reference and learns to correct its errors.
\begin{figure}[!b]
    \centering
    \begin{tikzpicture}
        \node[anchor=south west,inner sep=0] (img) at (0,0)
        {\includegraphics[width=0.98\textwidth]{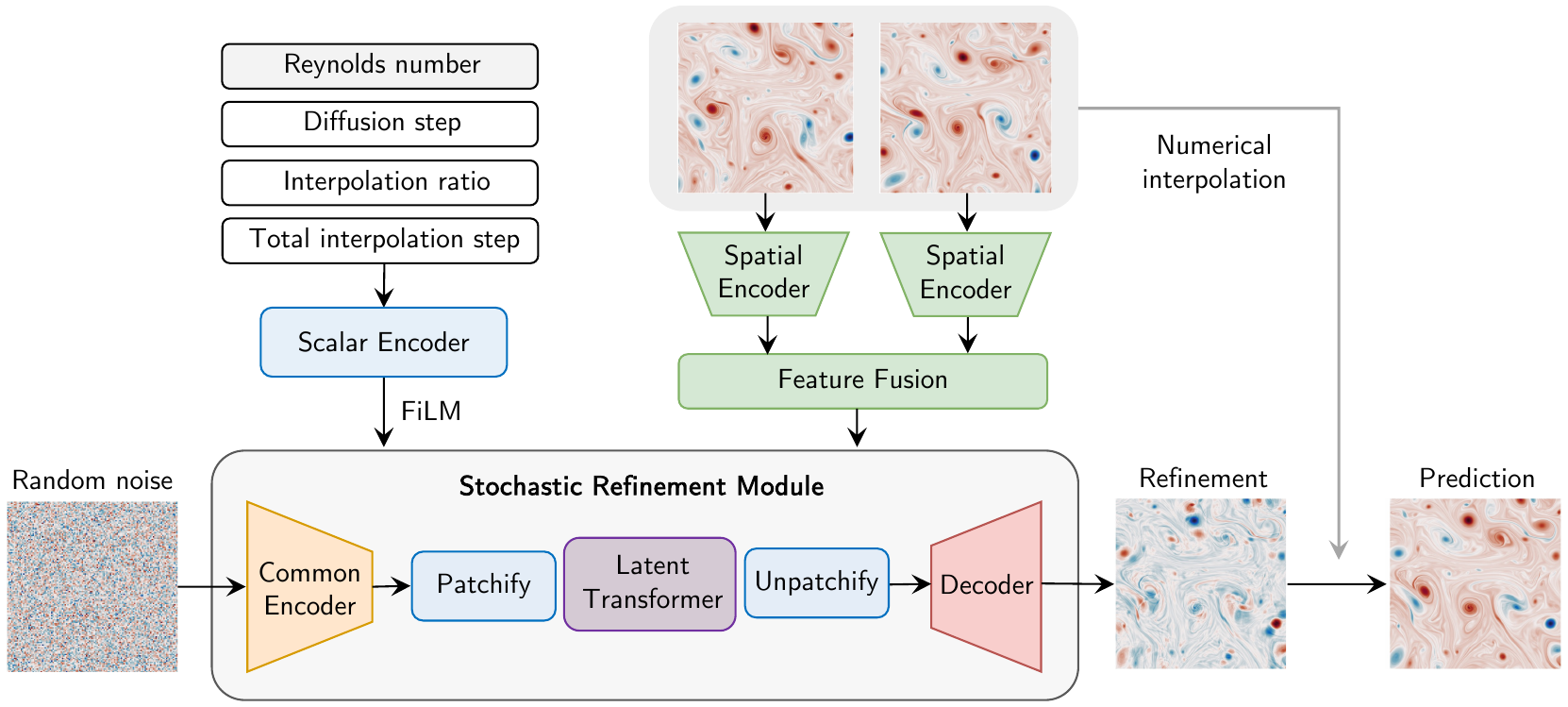}};
        \begin{scope}[x={(img.south east)},y={(img.north west)}]
            \fill[white] (0.152,0.632) rectangle (0.337,0.690);
            \node[font=\scriptsize] at (0.245,0.661) {Endpoint separation};
            \fill[white] (0.188,0.798) rectangle (0.300,0.854);
            \node[font=\scriptsize] at (0.244,0.825) {Diffusion time};
            \fill[white] (0.724,0.716) rectangle (0.825,0.822);
            \node[font=\scriptsize,align=center] at (0.775,0.769) {Linear\\interpolation};
            \fill[white] (0.721,0.292) rectangle (0.814,0.351);
            \node[font=\scriptsize] at (0.768,0.321) {Residual};
        \end{scope}
    \end{tikzpicture}
    \put(-240, 170){$\mathbf{u}_0$}
    \put(-180, 170){$\mathbf{u}_1$}
    \put(-435, 3){$\mathbf{z}_n$}
    \put(-110, 3){$\widehat{\mathbf{r}}_\tau$}
    \put(-30, 2){$\widehat{\mathbf{u}}_\tau$}
    \put(-73, 32){\scalebox{1.6}{$\mathbf{\oplus}$}}
\caption{
Overview of \model{} for temporal interpolation between two observed physical fields. Linear interpolation provides an initial estimate of the intermediate state, while a conditional diffusion model generates the correction needed to reconstruct the missing field. The two endpoint observations are processed by a shared spatial encoder, and their features are combined and supplied to the denoising network through learned gates. The requested interpolation time, observation interval, available physical parameters, and diffusion noise level are incorporated through feature-wise linear modulation (FiLM). After the final denoising step, the generated residual is added to the linear estimate to obtain the reconstructed intermediate field.
}\label{fig:overview_model}
\end{figure}

\section{Method}\label{sec:method}

In this section, we describe how \model{} reconstructs a physical field at a requested time between two observations. Our approach combines linear interpolation with a conditional diffusion model. The linear interpolation provides an initial estimate of the missing field, while the diffusion model learns to correct its errors using information from both observations. We first explain how this correction is constructed and learned, and then describe how the observed fields, temporal variables, and physical parameters enter the denoising network. Finally, we discuss how training examples are selected and how intermediate fields are reconstructed at inference time. An overview of the method is shown in Figure~\ref{fig:overview_model}.

\subsection{Linear Interpolation and Residual Learning}
\label{s:coarse_interp}

Given two observed fields $\mathbf{u}_0$ and $\mathbf{u}_1$, we begin by estimating the missing field through linear interpolation. For a requested intermediate time specified by $\tau$, we compute
\begin{equation}
    \label{eq:linear_interp}
    \widetilde{\mathbf{u}}_\tau
    =
    (1-\tau)\mathbf{u}_0+\tau\mathbf{u}_1.
\end{equation}
The interpolation is performed independently at each grid point, with the relative contribution of each observation determined by $\tau$. At the midpoint, both observations receive equal weight. As the requested time approaches either endpoint, the corresponding observation contributes more strongly to the estimate. This construction is computationally inexpensive, introduces no learned parameters, and exactly reproduces the observed fields at the two endpoints. It also requires no knowledge of the governing equations, making it applicable to both simulated and measured physical fields.

However, linear interpolation does not explicitly account for the movement or deformation of structures between observations. Consider a vortex that changes position over the observation interval. At its initial location, the first field may contain large vorticity values, while the second field contains smaller values because the vortex has moved. Averaging the two fields at fixed grid points can therefore produce weakened copies of the vortex near its observed locations, rather than placing it at its intermediate position. The problem becomes more pronounced if the vortex also stretches or changes its vorticity distribution. Similar difficulties arise in precipitation data, where a rain cell may move while its intensity increases or decreases. These examples illustrate why even an inexpensive and well-defined numerical interpolation can provide an inaccurate estimate of the intermediate state.

Rather than generating the entire intermediate field directly, we train a model to correct the errors in this initial estimate. During training, the intermediate field $\mathbf{u}_\tau$ is known, allowing us to compute the difference between the observed field and its linear interpolation,
\begin{equation}
    \label{eq:refinement}
    \mathbf{r}_\tau
    =
    \mathbf{u}_\tau-\widetilde{\mathbf{u}}_\tau.
\end{equation}
We refer to this difference as the residual. It describes the correction required at every grid point to recover the intermediate state. Returning to the moving vortex, the residual contains negative corrections where the linear interpolation places too much vorticity and positive corrections where the intermediate vortex is missing. Adding these corrections can therefore change the location and shape of a structure, even though the residual is applied pointwise. The correction can also extend across larger regions of the field, since linear interpolation in time does not separate spatial scales.

The motivation for this formulation is that the two observations already provide a useful starting point. Their weighted average contains information about the field values at both ends of the interval, even when the resulting spatial structures are inaccurate. Learning the residual allows the model to concentrate on the errors of this estimate. However, the residual need not be small or simpler than the original field, particularly when substantial changes occur between observations. We therefore treat linear interpolation as a convenient numerical reference whose usefulness
must be established empirically. In Section~\ref{sec:ablation}, we compare residual learning with direct generation of the intermediate field under matched model and training settings.

\subsection{Learning the Residual with Diffusion}

The residual is available during training because the intermediate field is known. At inference time, however, we only have the endpoint observations and the conditions associated with the requested reconstruction. The task is therefore to estimate a residual that is consistent with this information. We use a conditional diffusion model~\citep{ho2020denoising,song2021ddim}, building on the residual diffusion approach introduced in FLEX~\citep{erichson2025flex}. Diffusion models learn to recover data from progressively corrupted versions, allowing them to generate fields from noise while using the available observations as conditions. This is useful for our setting because the residual may contain complex spatial structures that are difficult to describe through a prescribed motion model. Our objective is accurate intermediate-state reconstruction; we do not assume that the generated samples provide calibrated estimates of physical uncertainty.

To train the diffusion model, we first compute the residual $\mathbf{r}_\tau$ from the known intermediate field. We then add Gaussian noise at a randomly selected diffusion time $n$,
\begin{equation}
    \label{eq:ddim_forward}
    \mathbf{z}_n
    =
    \alpha_n\mathbf{r}_\tau
    +
    \sigma_n\boldsymbol{\epsilon},
\end{equation}
where $\boldsymbol{\epsilon}$ is sampled from a standard Gaussian distribution. The coefficients $\alpha_n$ and $\sigma_n$ determine how much of the original residual and sampled noise are present in $\mathbf{z}_n$. We use a cosine log-signal-to-noise ratio schedule with $\alpha_n^2+\sigma_n^2=1$. At low noise levels, the corrupted field still resembles the original residual, while at high noise levels most of its structure is obscured. By training on different noise levels, the network learns to recover information from both strongly and weakly corrupted inputs.
Thereby, it is important to distinguish diffusion time from the physical time of the reconstructed field. The interpolation ratio $\tau$ identifies the physical state we wish to reconstruct, whereas $n$ controls the amount of artificial noise added during diffusion. For example, when reconstructing the midpoint between two observations, $\tau=0.5$ remains fixed throughout the denoising process. Only the diffusion noise level changes as the model approaches its final residual estimate.

Following the velocity-prediction parameterization used in diffusion models, we define the training target
\begin{equation}
    \label{eq:velocity_target}
    \mathbf{v}_n
    =
    \alpha_n\boldsymbol{\epsilon}
    -
    \sigma_n\mathbf{r}_\tau.
\end{equation}
The term velocity refers to the diffusion parameterization and is unrelated to the physical velocity of a fluid. It combines the clean residual and the sampled noise according to the current noise level. This parameterization is convenient because a prediction of $\mathbf{v}_n$ can be converted directly into an estimate of the clean residual, as we show in the inference procedure below.

We train a denoising network $v_\theta$ to predict $\mathbf{v}_n$ from the noisy residual and the available conditions. The network receives the endpoint fields $\mathbf{u}_0$ and $\mathbf{u}_1$, the temporal variables $\mathbf{p}=(\tau,\Delta t)$, available physical parameters $\boldsymbol{\lambda}$, and the diffusion time $n$. The training objective is
\begin{equation}
    \label{eq:denoising_loss}
    \mathcal{L}_{\mathrm{denoise}}(\theta)
    =
    \mathbb{E}
    \left[
    \left\|
    v_\theta(
        \mathbf{z}_n,n;
        \mathbf{u}_0,\mathbf{u}_1,
        \mathbf{p},\boldsymbol{\lambda}
    )
    -
    \mathbf{v}_n
    \right\|_1
    \right].
\end{equation}
The expectation is taken over training examples, randomly sampled diffusion times, and Gaussian noise. We use the $\ell_1$ loss to penalize errors in the predicted diffusion velocity. Since the training residual is known, the target velocity can be computed directly for every sampled noise level. The network learns to predict this quantity while using the endpoint observations and physical conditions to identify the intermediate state being reconstructed.
For the denoising network, we adopt the encoder-transformer-decoder architecture of
FLEX~\citep{erichson2025flex}. This architecture combines convolutional processing of local spatial structures with attention at lower spatial resolutions. The combination is suitable for physical fields that contain both localized features and structures extending over larger regions of the domain. We retain this backbone and adapt its conditioning to use information from both observed endpoints, as described next.

\subsection{Conditioning on Observations and Physical Variables}
\label{s:structured_condition}

Accurate reconstruction depends on more than the noisy residual supplied to the denoiser. The network also needs information about the spatial structures in the
two observations and the physical conditions under which they evolve. These inputs play different roles. The endpoint fields describe where structures are located and how their values differ, while the interpolation ratio, observation interval, and
physical parameters describe the requested reconstruction. We therefore incorporate the observed fields through spatial feature maps and the scalar variables through feature-wise modulation.

\paragraph{Spatial conditioning.}
Although the linear estimate is constructed from both observations, averaging them removes some information about their individual spatial structures. For example, exchanging the initial and final fields leaves the midpoint interpolation unchanged, even though it reverses the observed change over time. The denoising network therefore receives both endpoint fields separately, allowing it to distinguish the beginning and end of the interval.

We process the endpoint fields with a shared spatial encoder $E_\psi$. At resolution level $\ell$, the encoder produces the feature maps $\mathbf{f}_0^{(\ell)}=E_\psi^{(\ell)}(\mathbf{u}_0)$ and $\mathbf{f}_1^{(\ell)}=E_\psi^{(\ell)}(\mathbf{u}_1)$. Using the same encoder for both observations allows corresponding features to be computed with the same learned filters. The resulting feature maps can then be compared and combined at each resolution.
We construct the spatial conditioning map by concatenating the endpoint features with their average and signed difference, followed by a learned projection
\begin{equation}
    \label{eq:spatial_condition}
    \mathbf{c}_{\mathrm{sp}}^{(\ell)}
    =
    \psi_\ell
    \left(
    \left[
    \mathbf{f}_0^{(\ell)},
    \mathbf{f}_1^{(\ell)},
    \frac{\mathbf{f}_0^{(\ell)}+\mathbf{f}_1^{(\ell)}}{2},
    \mathbf{f}_1^{(\ell)}-\mathbf{f}_0^{(\ell)}
    \right]
    \right).
\end{equation}
The brackets indicate concatenation along the feature channels, and $\psi_\ell$ denotes the learned projection. The individual endpoint features preserve information from each observation, while their signed difference explicitly represents how the encoded fields change across the interval. The average provides another combination of the same information. These operations introduce no new information beyond the original endpoint features, but they make several useful combinations available to the projection without requiring the network to construct them entirely through learned operations. We examine their contribution in the feature-fusion ablation.

The spatial conditioning maps are supplied to the denoising network at resolutions of $32\times32$ and $16\times16$ in the encoder and decoder. These lower-resolution features allow information from the endpoints to influence the reconstruction
at spatial scales extending beyond individual grid points. This is useful when the correction must account for the displacement or deformation of a structure across several locations. The subsequent network layers combine this information
with the noisy residual to estimate the required correction.

\paragraph{Scalar conditioning.}
The reconstruction also depends on quantities that describe the physical system and the requested time. The interpolation ratio $\tau$ specifies where the missing state lies within the observation interval, while $\Delta t$ determines the actual duration of that interval. These variables must be distinguished because the same interpolation ratio can correspond to very different amounts of physical evolution. Similarly, physical parameters such as the Reynolds number can change the dynamics
between observations. The diffusion time $n$ provides another condition, specifying the noise level at which the denoiser operates.

We collect these quantities into the scalar input $\mathbf{s}=[\boldsymbol{\lambda},\tau,\Delta t,n]$ and encode them using Fourier features followed by a multilayer perceptron. The resulting embedding is $\mathbf{e}_{\mathrm{sc}}=\mathrm{MLP}(\Phi(\mathbf{s}))$, where $\Phi$ denotes the Fourier feature mapping. The embedding is then projected to channel-wise scale and shift parameters at each denoising block. Following feature-wise linear modulation (FiLM)~\citep{perez2018film}, the hidden features are adjusted according to
\begin{equation}
    \label{eq:film}
    \widetilde{\mathbf{h}}^{(\ell)}
    =
    \left(1+\boldsymbol{\gamma}^{(\ell)}\right)
    \odot\mathbf{h}^{(\ell)}
    +
    \boldsymbol{\beta}^{(\ell)}.
\end{equation}
Here, $\mathbf{h}^{(\ell)}$ denotes the hidden feature map, and $\boldsymbol{\gamma}^{(\ell)}$ and $\boldsymbol{\beta}^{(\ell)}$ are the learned scale and shift parameters obtained from the scalar embedding. These parameters are broadcast across the spatial dimensions, so they adjust the contribution of entire feature channels according to the current conditions.

This provides a simple way for one denoising network to respond to different physical settings. For example, the network may need to make a larger correction when reconstructing the midpoint of a long observation interval than when the observations are close together. The scalar conditions allow the feature transformations to depend on these differences without requiring a separate model for each interval duration or physical parameter. FiLM is particularly convenient here because the same scalar conditions are relevant across the spatial domain, whereas the endpoint features retain information that varies from one location to another.

\paragraph{Combining spatial and scalar conditions.}
The two conditioning pathways meet inside the denoising network. After applying FiLM to the hidden features, we inject the spatial conditioning information through a learned gate
\begin{equation}
    \label{eq:spatial_injection}
    \mathbf{h}_{\mathrm{cond}}^{(\ell)}
    =
    \widetilde{\mathbf{h}}^{(\ell)}
    +
    \mathbf{g}^{(\ell)}
    \odot
    W_c^{(\ell)}
    \left(
    \mathbf{c}_{\mathrm{sp}}^{(\ell)}
    \right).
\end{equation}
The projection $W_c^{(\ell)}$ maps the spatial conditioning features to the required number of hidden channels. The gate is computed from the spatial conditioning map as $\mathbf{g}^{(\ell)}=\sigma(W_g^{(\ell)}(\mathbf{c}_{\mathrm{sp}}^{(\ell)}))$, where $\sigma$ is the sigmoid function. Its values determine how strongly the projected endpoint features contribute to the hidden representation. Smaller gate values reduce this contribution, while larger values allow more of the spatial information to enter the denoiser.
The gate is learned jointly with the rest of the network, allowing the strength of spatial conditioning to depend on the encoded observations. Together, the conditioning pathways give the denoiser access to the locations and changes of physical structures, the requested intermediate time, and the conditions under which the field evolves. The architecture thus combines spatially varying information from the endpoint fields with scalar information that describes the reconstruction as a whole.

\subsection{Training and Reconstruction at Arbitrary Times}
\label{s:training_inference}

Training examples are constructed from sequences containing known intermediate states. Each example consists of two endpoint observations and a target field between them. A straightforward approach would always select observations separated by the same number of recorded time steps. However, the reconstruction difficulty depends on the length of the observation interval, since physical structures generally have more time to move and change over longer intervals. We therefore vary the separation between endpoint observations during training, allowing the model to encounter different temporal gaps.

Let $T$ denote the number of intermediate states between the selected endpoints at the native temporal resolution of the dataset. The two observations are separated by $T+1$ native time intervals. If the native temporal spacing is $\Delta t_{\mathrm{data}}$, their physical separation is $\Delta t=(T+1)\Delta t_{\mathrm{data}}$. During training, we sample $T$ from a dataset-specific range and then select a target index $k$ uniformly from $\{1,\ldots,T\}$. The interpolation ratio is $\tau=k/(T+1)$.
For example, choosing $T=20$ gives 20 intermediate states between two endpoint observations, which are separated by 21 native time intervals. If the target is the seventh intermediate state, then $\tau=7/21=1/3$. This example illustrates why the number of missing states and the physical duration of the interval are related but distinct. The former depends on the sampling frequency of the dataset, while the latter determines the time over which the underlying system evolves. The training procedure supplies both the relative target position and the physical interval duration to the model. The dataset-specific sampling ranges and implementation settings are given in Appendix~\ref{app:model_config}.

For each selected training triplet, we compute the linear interpolation from the two endpoint fields and subtract it from the known intermediate state to obtain the residual. We then sample Gaussian noise and a diffusion time, construct the noisy residual, and train the denoising network to predict the corresponding diffusion velocity. The endpoint features and scalar conditions identify the reconstruction problem, while the sampled noise level determines how much of the residual must be recovered. 

At inference time, the intermediate field is unknown. Given the two observations, the requested target time, and the available physical parameters, we first compute the linear estimate $\widetilde{\mathbf{u}}_\tau$. The endpoint observations are also processed by the shared encoder to obtain the spatial conditioning maps. Residual generation begins from Gaussian noise and proceeds through a DDIM-style sampling process~\citep{song2021ddim}. We use three denoising steps, with the physical target time and endpoint conditions held fixed throughout the process.

At each step, the denoising network predicts the diffusion velocity $\widehat{\mathbf{v}}_n$ from the current noisy residual and its conditions. We convert this prediction into an estimate of the clean residual using
\begin{equation}
    \label{eq:recover_residual}
    \widehat{\mathbf{r}}_{\tau,n}
    =
    \alpha_n\mathbf{z}_n
    -
    \sigma_n\widehat{\mathbf{v}}_n.
\end{equation}
This relation follows directly from the noise construction and the velocity-prediction target. If the network predicts the exact target velocity, substituting the definitions of $\mathbf{z}_n$ and $\mathbf{v}_n$ cancels the noise and recovers $\mathbf{r}_\tau$, since $\alpha_n^2+\sigma_n^2=1$. The predicted velocity therefore provides an estimate of the clean correction at each noise level. The sampler uses this estimate to update the noisy residual before the next denoising step.
After the final step, the model produces the residual estimate $\widehat{\mathbf{r}}_\tau$. The missing intermediate field is reconstructed by adding this correction to the original linear estimate
\begin{equation}
    \label{eq:final_reconstruction}
    \widehat{\mathbf{u}}_\tau
    =
    \widetilde{\mathbf{u}}_\tau
    +
    \widehat{\mathbf{r}}_\tau.
\end{equation}
The same numerical reference is used to define the training residual and to reconstruct the intermediate field. Consequently, the error in the reconstructed field is equal to the error in the generated residual. If the model predicts the residual exactly, the intermediate field is recovered regardless of the accuracy of the linear estimate. In practice, the quality of the reconstruction depends on how well the model learns this correction.

Because the requested target time is included in the conditioning variables, the trained model can reconstruct fields at different positions within an observation interval. Several target times can also be processed together by placing them in the same sampling batch. Each intermediate field is generated independently, so the procedure does not explicitly enforce temporal consistency between consecutive reconstructions. 
\section{Experiments}\label{sec:experiments}

In this section, we evaluate how accurately \model{} reconstructs intermediate states across different physical systems. We begin with simulated turbulent flows, where the governing equations and physical parameters are known, and then consider radar precipitation and sea surface temperature observations. These datasets allow us to examine interpolation in settings with different spatial structures and rates of temporal change. We compare \model{} with numerical interpolation, supervised reconstruction, and methods developed for video frame interpolation. We then examine how reconstruction errors vary within an observation interval, how the model performs at different temporal gaps, and which design choices contribute to its accuracy.

\subsection{Reconstruction Across Different Physical Systems}
\label{s:results}

\textbf{Evaluation setup.}
We evaluate reconstruction quality using relative Frobenius norm error (RFNE), Pearson correlation (PC), and structural similarity index (SSIM)~\citep{wang2004image}. RFNE measures the relative difference between the reconstructed and observed fields, while PC measures their linear correlation and SSIM compares their spatial structure. These metrics provide complementary information. For example, a reconstruction may reproduce the spatial pattern of a flow while underestimating its vorticity values, resulting in high correlation but appreciable reconstruction error. RFNE and PC are computed using fields in their original physical units, whereas SSIM is computed after normalization. The precise metric definitions are given in Appendix~\ref{app:eval_metrics}. We report results from the implemented stochastic sampling procedure and use the same evaluation metrics for all methods.

Our comparisons include linear interpolation, which provides the numerical reference used by \model{}, and a supervised U-Net~\citep{ronneberger2015u} that predicts intermediate fields directly. We also evaluate Super SloMo~\citep{jiang2018super}, which reconstructs intermediate images using optical flow, and EDEN~\citep{zhang2025eden}, a diffusion-based video frame interpolation model. These methods represent different ways of using the observed endpoints. Linear interpolation averages their values, optical-flow methods estimate how image content moves, and the learned models predict or generate intermediate fields from the available observations. We adapt the models to the scientific datasets and compare their reconstructions with the same ground-truth intermediate states. Additional implementation details and a supervised PIT-UNet comparison are provided in Appendix~\ref{app:model_config} and Appendix~\ref{app:pit_unet}.

\textbf{Navier--Stokes Kraichnan turbulence.} We first consider the Navier--Stokes Kraichnan turbulence (NSKT) dataset, which extends the simulations used in SuperBench~\citep{ren2025superbench}. The data consist of vorticity fields from two-dimensional incompressible fluid simulations. Turbulence provides a useful test because vortices move, stretch, and interact over time, making their intermediate states difficult to reconstruct from widely separated observations. The Reynolds number $Re$ controls the relative importance of inertial and viscous effects, allowing us to examine the model under different flow conditions.
The simulations are generated on a $2048\times2048$ grid, from which $256\times256$ spatial crops are extracted for training and evaluation. Models are trained using $Re\in\{2000,4000,8000,16000,32000\}$. We evaluate them at Reynolds numbers included in training, at unseen values within the training range, and at values outside that range. The main comparison in Table~\ref{tab:ns_re_eval} focuses on $Re=12000$, $24000$, and $36000$, using 20 intermediate states between the observed endpoints. The first two Reynolds numbers lie within the range spanned by training, whereas $Re=36000$ is larger than every training value.

\begin{table*}[t]
\centering
\caption{
\model{} achieves the lowest reconstruction error at all three unseen Reynolds numbers. All methods are trained using $Re\in\{2000,4000,8000,16000,32000\}$ and evaluated with
20 intermediate states between the endpoint observations. The values $Re=12000$ and $24000$ are unseen but lie within the training range, while $Re=36000$ tests extrapolation beyond that range. At $Re=36000$, \model{} achieves $10.48\%$ RFNE, compared with $22.87\%$ for Super SloMo
and $25.52\%$ for EDEN. RFNE denotes relative Frobenius norm error, PC denotes Pearson correlation, and SSIM denotes structural similarity. Inference time (IT) is reported in milliseconds per generated intermediate field. Bold values indicate the best reconstruction scores.
}
\label{tab:ns_re_eval}
\resizebox{0.9\linewidth}{!}{
\begin{tabular}{lcc ccc ccc ccc}
\toprule
& & &
\multicolumn{3}{c}{$Re=12000$} &
\multicolumn{3}{c}{$Re=24000$} &
\multicolumn{3}{c}{$Re=36000$}
\\
\cmidrule(lr){4-6}
\cmidrule(lr){7-9}
\cmidrule(lr){10-12}
Method
& \#Params [M]
& IT [ms]
& RFNE $\downarrow$
& PC $\uparrow$
& SSIM $\uparrow$
& RFNE $\downarrow$
& PC $\uparrow$
& SSIM $\uparrow$
& RFNE $\downarrow$
& PC $\uparrow$
& SSIM $\uparrow$
\\
\midrule
Linear Interp.
& --
& 0.37
& 32.12\% & 0.93 & 0.43
& 39.55\% & 0.90 & 0.35
& 43.96\% & 0.88 & 0.30
\\
U-Net
& 260.33
& 29.41
& 39.54\% & 0.92 & 0.41
& 50.97\% & 0.88 & 0.27
& 59.94\% & 0.83 & 0.18
\\
EDEN
& 194.00
& 9.53
& 13.15\% & 0.99 & 0.82
& 20.14\% & 0.98 & 0.71
& 25.52\% & 0.96 & 0.63
\\
Super SloMo
& 39.59
& 3.54
& 13.20\% & 0.99 & 0.83
& 18.38\% & 0.98 & 0.74
& 22.87\% & 0.97 & 0.67
\\
\midrule
\rowcolor{orange!12}
\model{}
& 110.4
& 35.39
& \textbf{4.11\%} & \textbf{1.00} & \textbf{0.96}
& \textbf{7.10\%} & \textbf{1.00} & \textbf{0.92}
& \textbf{10.48\%} & \textbf{0.99} & \textbf{0.87}
\\
\bottomrule
\end{tabular}
}
\end{table*}

Table~\ref{tab:ns_re_eval} shows that \model{} achieves lower reconstruction error than all evaluated baselines at the three Reynolds numbers. At $Re=12000$, its RFNE is $4.11\%$, compared with $13.15\%$ for EDEN and $13.20\%$ for Super SloMo. The differences become larger at $Re=36000$, where \model{} obtains $10.48\%$ RFNE, compared with $22.87\%$ for Super SloMo and $25.52\%$ for EDEN. The improvements in SSIM are also substantial, indicating that the reconstructed fields more closely reproduce the spatial patterns of the observed vorticity. The qualitative comparison in Figure~\ref{fig:nskt_re36000_snapshot_comparison} illustrates these differences in vortex shape and location.
The comparison with linear interpolation is particularly informative because it measures how much the learned correction improves the initial numerical estimate. At $Re=36000$, linear interpolation has an RFNE of $43.96\%$, compared with $10.48\%$ for \model{}. This large difference shows that the linear reference is inaccurate on its own for this turbulent regime, but the learned correction can substantially improve the reconstructed field. The comparison with Super SloMo also indicates that estimating motion between the observations does not fully resolve the reconstruction problem in these flows. However, comparisons between complete models do not isolate the effect of any individual architectural component. We examine the specific contribution of residual learning in Section~\ref{sec:ablation}.

The complete evaluation across Reynolds numbers is reported in Appendix~\ref{app:nskt_complete_re}. It includes Reynolds numbers seen during training, unseen values within the training range, and extrapolation below and above that range. \model{} achieves the lowest RFNE among the evaluated methods at every reported Reynolds number. These results show that the trained model can reconstruct fields at physical parameter values not included in its training set. Since the main sweep uses one held-out trajectory per Reynolds number, it does not by itself establish how reconstruction accuracy varies across independent realizations of the same flow.

We therefore also evaluate a fixed trained checkpoint on additional trajectories at $Re=8000$, $12000$, and $32000$, as described in Appendix~\ref{sec:trajectory_replication}. The reported variation is relatively small for $Re=8000$ and $12000$, but substantially larger for $Re=32000$. This indicates that reconstruction difficulty can depend on the particular flow realization, even when the Reynolds number is fixed.

\textbf{Shanghai radar precipitation.} We next evaluate \model{} on the Shanghai radar dataset ~\citep{chen2020deep}, following the preprocessing procedure of~\citet{yu2024diffcast}. The observations are recorded at approximately six-minute intervals and cover a $501\,\mathrm{km}\times501\,\mathrm{km}$ region on a $128\times128$ grid. We reconstruct precipitation fields at eight intermediate times between the selected observations.
Radar precipitation differs from the simulated vorticity fields in several ways. Precipitation is spatially intermittent, and its local intensity can change rapidly. Rain cells may move, intensify, weaken, or disappear between observations. Consequently, a method that estimates their displacement may still produce incorrect precipitation values. Small errors in the position of intense rainfall can also contribute substantially to the relative reconstruction error.

\begin{table*}[t]
\centering
\caption{
\model{} achieves the lowest reconstruction error on
Shanghai radar precipitation and sea surface temperature. The evaluations use eight intermediate states for radar and ten for sea surface temperature. Radar precipitation shows substantially larger reconstruction errors, while linear interpolation is already accurate for the more slowly varying temperature fields. On SST, \model{} reduces RFNE from $1.94\%$ for linear interpolation to $1.81\%$. Inference time (IT) is reported in milliseconds per generated intermediate field. Bold values indicate the best reconstruction scores.
}
\label{tab:sh_sst_eval}
\resizebox{0.85\linewidth}{!}{
\begin{tabular}{lccccc cccc}
\toprule
& &
\multicolumn{4}{c}{Shanghai Radar Data}
& \multicolumn{4}{c}{Sea Surface Temperature} \\
\cmidrule(lr){3-6}
\cmidrule(lr){7-10}
Method
& \#Params [M]
& IT [ms]
& RFNE $\downarrow$
& PC $\uparrow$
& SSIM $\uparrow$
& IT [ms]
& RFNE $\downarrow$
& PC $\uparrow$
& SSIM $\uparrow$
\\
\midrule
Linear Interp.
& --
& 0.09 & 47.78\% & 0.84 & 0.79
& 0.12 & 1.94\% & 0.99 & 0.83
\\
U-Net
& 260.33
& 13.25 & 48.51\% & 0.85 & 0.82
& 13.27 & 2.29\% & 0.996 & 0.80
\\
EDEN
& 194.00
& 2.32 & 56.54\% & 0.77 & 0.77
& 2.45 & 2.48\% & 0.996 & 0.77
\\
Super SloMo
& 39.59
& 1.45 & 36.77\% & 0.91 & 0.87
& 1.45 & 2.28\% & 0.996 & 0.79
\\
\midrule
\rowcolor{orange!12}
\model{}
& 110.4
& 10.53 & \textbf{33.52\%} & \textbf{0.92}
& \textbf{0.87}
& 9.35 & \textbf{1.81\%} & \textbf{0.998}
& \textbf{0.87}
\\
\bottomrule
\end{tabular}
}
\end{table*}

Table~\ref{tab:sh_sst_eval} shows that \model{} achieves an RFNE of $33.52\%$ on radar precipitation, compared with $36.77\%$ for Super SloMo and $47.78\%$ for linear interpolation. The improvement over Super SloMo is smaller than on the turbulence dataset, but \model{} also achieves the highest correlation and structural similarity. The remaining error is appreciable, reflecting the difficulty of reconstructing spatially localized precipitation from temporally separated observations. In particular, neither endpoint necessarily reveals the precise timing of a short-lived precipitation event.
Figure~\ref{fig:shanghai_snapshot_comparison} provides qualitative examples of the reconstructed precipitation fields. These examples help illustrate differences in the location and intensity of rainfall, while the metrics in Table~\ref{tab:sh_sst_eval} measure reconstruction accuracy over the evaluation set. We further examine how radar reconstruction changes with the temporal gap in Appendix~\ref{app:temporal_gap}.

\paragraph{Sea surface temperature.}
Our third dataset consists of sea surface temperature (SST) fields used in HelmFluid~\citep{xing2023helmfluid}, derived from Copernicus Marine Service data ~\citep{ECMWF_MDS_2023}. We use a spatial resolution of $128\times128$ and daily observations, with ten intermediate states between the selected endpoints. Models are trained on observations from the South Atlantic, Indian Ocean, and North Pacific regions, and evaluated on the held-out South Pacific region. This experiment examines whether the learned interpolation method can be applied to an ocean region not represented in training.
SST provides a useful contrast to radar precipitation and turbulence. Temperature fields often contain large, smoothly varying spatial structures, and their changes between nearby observations can be relatively small. As a result, linear interpolation already provides an accurate estimate of the missing state. Table~\ref{tab:sh_sst_eval} shows that its RFNE is only $1.94\%$, substantially below the corresponding errors on the other datasets. This makes SST a more demanding test of whether a learned correction provides an improvement over a strong numerical reference.

On this dataset, \model{} reduces RFNE from $1.94\%$ for linear interpolation to $1.81\%$,
while increasing SSIM from $0.83$ to $0.87$. The absolute reduction in RFNE is modest, as expected when the initial interpolation is already accurate. The improvement in SSIM indicates better agreement in the spatial structure of the reconstructed temperature fields. Figure~\ref{fig:sst_snapshot_comparison} shows qualitative examples from the held-out region. Together with the radar and turbulence results, this experiment shows that the residual formulation can improve reconstruction both when linear interpolation has large errors and when those errors are already small.

\subsection{How Reconstruction Error Changes Within an Interval}
\label{s:analysis}

The preceding comparisons summarize reconstruction accuracy over the evaluated intermediate states. However, errors need not be uniform throughout an observation interval. A target near the first endpoint is close in time to an observed field, whereas a target near the midpoint is farther from both observations. The latter generally requires reconstructing more of the intervening evolution. We therefore examine how reconstruction error changes with the requested target time.

\begin{figure*}[t]
    \centering
    \includegraphics[width=\textwidth]{
        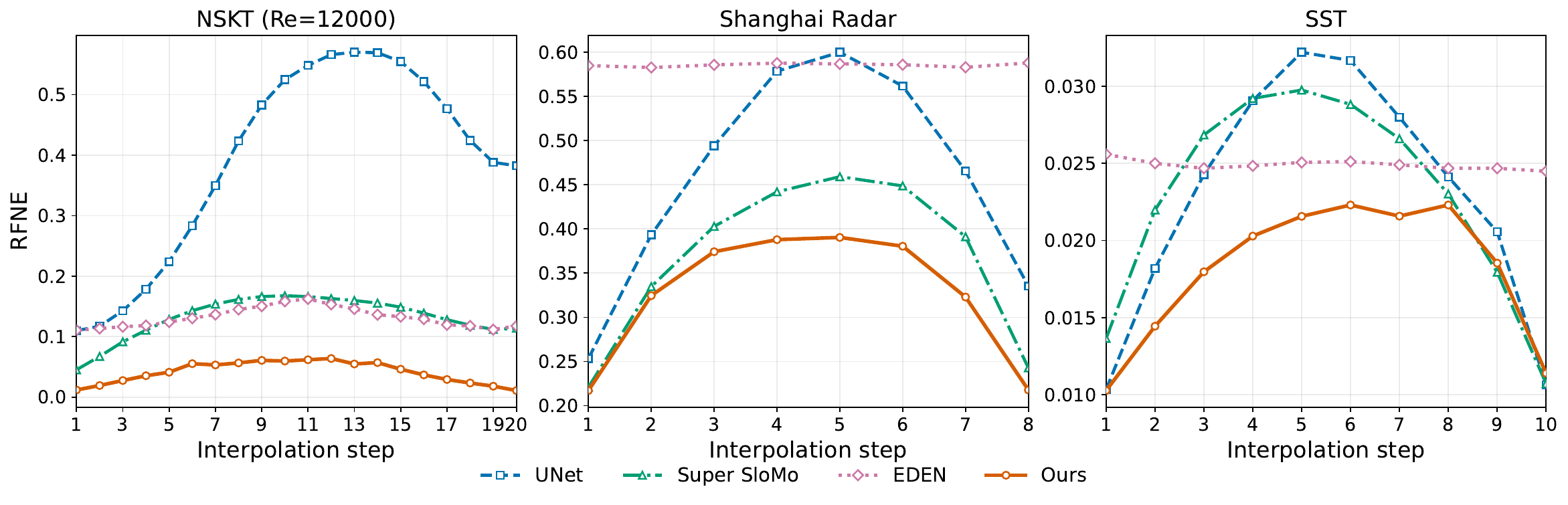
    }
    \caption{
    \model{} achieves lower reconstruction error across
the observation interval on NSKT, Shanghai radar, and SST. The panels report RFNE at each requested intermediate step, using 20 intermediate states for NSKT at
    $Re=12000$, eight for radar, and ten for SST.
Errors are generally largest near the middle of the interval, where the requested field is farthest in time from both observations. The curves display RFNE as a fraction rather than a percentage.
    }
    \label{fig:error_steps_analysis}
\end{figure*}

Figure~\ref{fig:error_steps_analysis} reports RFNE at each intermediate step for NSKT at
$Re=12000$, Shanghai radar precipitation, and
SST. The error curves show a similar pattern for several methods. Errors tend to increase as the requested time moves away from the first observation and decrease again as it approaches the second. The largest errors often occur near the midpoint, where the requested field is farthest in time from either observed state. The curves are not necessarily symmetric, since physical structures may change at different rates within the interval.
The differences between methods are particularly visible on NSKT. The U-Net error increases strongly toward the middle of the interval, whereas \model{} maintains a substantially smaller error throughout. Super SloMo and EDEN reduce the error relative to U-Net, but remain above \model{} over most of the interval. Radar shows a similar difficulty near the midpoint, although the errors are larger for every method. On SST, the error curves are closer together because the underlying temporal changes are smaller. Nevertheless, \model{} maintains the lowest error across the evaluated intermediate steps.

These results help explain why interpolation accuracy depends on the position of the requested time. The endpoint observations are most informative near their own times, while the model must account for more unobserved evolution near the center of the interval. The residual formulation improves reconstruction throughout this interval, including the intermediate times where the task is most difficult.

\subsection{Spatial Structure, Spectral Content, and Inference Cost}

Reconstruction error measures how closely the predicted field matches the target, but it does not describe every aspect of the reconstructed spatial structure. This is particularly relevant for turbulence, where vortices and their smaller features contribute to the appearance and statistics of the flow. We therefore complement the pointwise evaluation with a comparison of spatial spectra. We also examine inference cost, since reconstructing many missing snapshots can require a substantial number of model evaluations.

\begin{figure*}[t]
    \centering
    \begin{subfigure}[t]{0.45\textwidth}
        \centering
        \includegraphics[width=\linewidth]{
            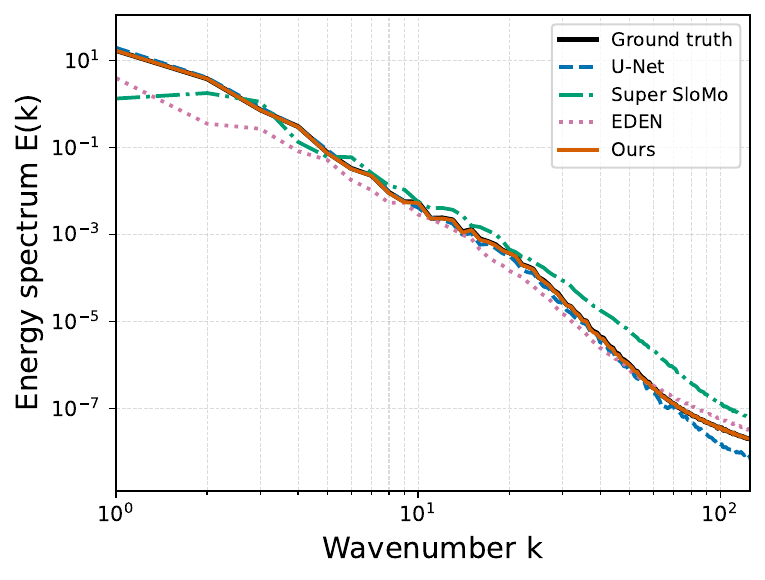
        }
        \caption{Windowed vorticity power spectrum}
        \label{fig:energy_spectrum}
    \end{subfigure}
    \hfill
    \begin{subfigure}[t]{0.45\textwidth}
        \centering
        \includegraphics[width=\linewidth]{
            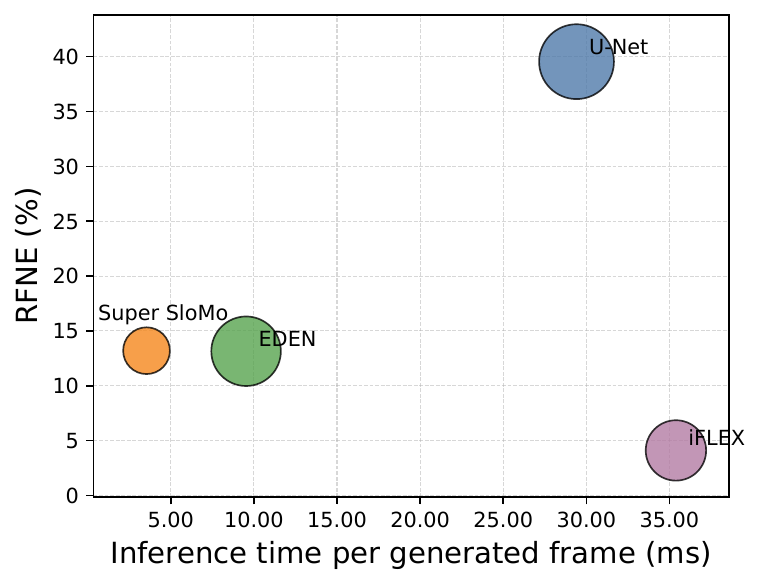
        }
        \caption{Reconstruction error and inference cost}
        \label{fig:efficiency_analysis}
    \end{subfigure}
    \caption{
Reconstruction accuracy depends on both spatial structure and computational cost. (a) Comparison of the ground-truth and reconstructed spatial spectra for NSKT at $Re=12000$. The intended diagnostic is the windowed vorticity power spectrum; the underlying figure export requires the verification noted in the text. (b) RFNE versus inference time for the evaluated methods at the same Reynolds number. \model{} achieves $4.11\%$ RFNE but requires $35.39\,\mathrm{ms}$ per generated field, compared with $13.20\%$ and $3.54\,\mathrm{ms}$ for Super SloMo. Bubble size represents model parameter count.
    }
    \label{fig:analysis_summary}
\end{figure*}

\textbf{Spectral comparison.}
Figure~\ref{fig:analysis_summary}(a) compares the spectra of reconstructed and ground-truth NSKT fields at $Re=12000$. The intended diagnostic is the windowed vorticity power spectrum, which measures how the spatial variation of vorticity is distributed across wavenumbers. This differs from a kinetic-energy spectrum and should not be interpreted as a direct measure of energy conservation. The spectrum is computed from spatial crops, which are windowed to reduce artifacts from their nonperiodic boundaries. The calculation is described in Appendix~\ref{app:eval_metrics}.

\textbf{Inference cost.}
Figure~\ref{fig:analysis_summary}(b) compares RFNE with inference time and model size for NSKT at $Re=12000$. \model{} obtains lower reconstruction error than the other learned methods, but this improvement comes at a higher inference cost. It requires $35.39\,\mathrm{ms}$ per generated field, compared with $3.54\,\mathrm{ms}$ for Super SloMo and $9.53\,\mathrm{ms}$ for EDEN. The difference is relevant when many intermediate fields must be reconstructed from long simulation records.
The number of network evaluations also matters. Although \model{} uses only three diffusion sampling steps, it must evaluate the denoiser at each step for every requested intermediate state. Several target times can be processed in one batch, but doing so does not reduce the sampling procedure to a single network evaluation. The reported latency therefore provides an important complement to the accuracy measurements. For applications where a coarse reconstruction is sufficient, linear interpolation or a faster learned method may remain preferable.

\subsection{Reconstruction Across Different Temporal Gaps}

The temporal separation between observations is another factor that affects reconstruction difficulty. Even when the requested target has the same relative position within the interval, a longer interval gives the physical system more time to evolve. This distinction motivated conditioning \model{} on both the interpolation ratio $\tau$ and the actual interval duration
$\Delta t$. We now examine how reconstruction accuracy changes when the number of intermediate states differs from those used during training.

\begin{figure*}[t]
    \centering
    \includegraphics[width=0.98\textwidth]{
        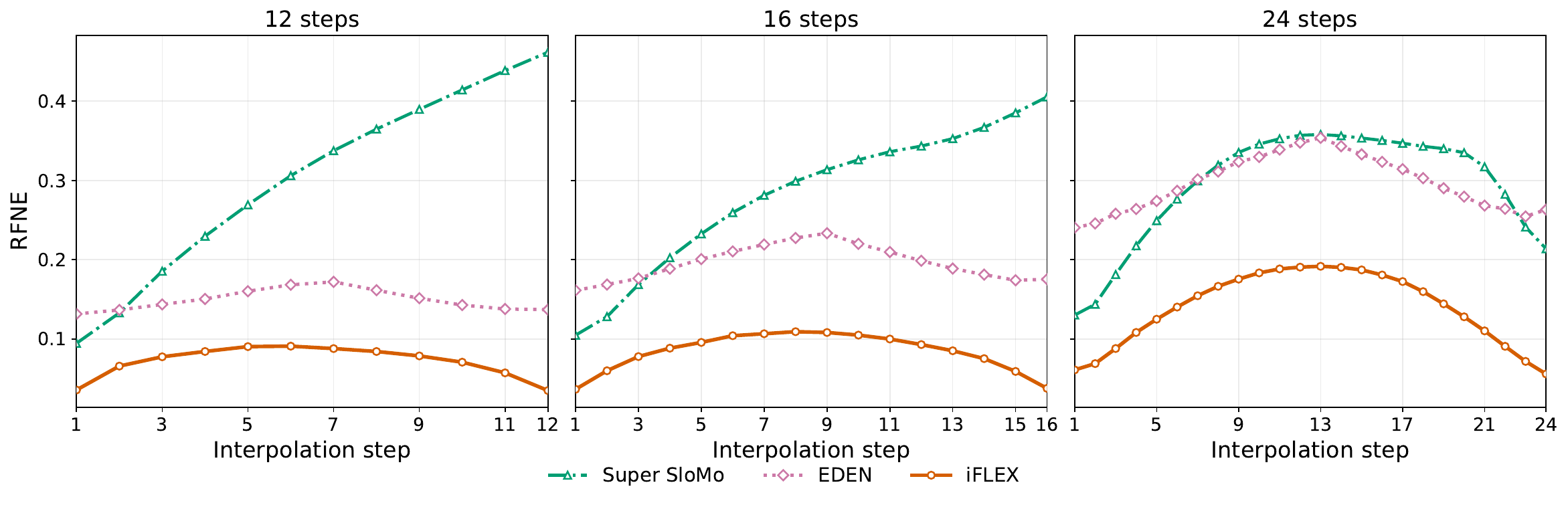
    }
    \caption{
    \model{} achieves lower reconstruction error than Super SloMo and EDEN across three temporal gaps on NSKT at $Re=36000$. The panels show RFNE at each intermediate step for $T=12$, $16$, and $24$. During training, \model{} samples $T\in\{5,\ldots,20\}$; therefore, the first two gaps lie within its training range, while $T=24$ exceeds it. Errors generally increase toward the midpoint of the observation interval, particularly for the longer gap. RFNE is plotted as a fraction.
    }
    \label{fig:interp_step_generalization}
\end{figure*}

Figure~\ref{fig:interp_step_generalization} shows results on NSKT at $Re=36000$ for $T=12$, $16$, and $24$ intermediate states. During training, \model{} samples the number of intermediate states from $T\in\{5,\ldots,20\}$. Thus, $T=12$ and $T=16$ lie within the training range, whereas $T=24$ evaluates a longer interval than those used for training. The figure compares \model{} with Super SloMo and EDEN at each requested intermediate step.

\model{} achieves lower RFNE than both baselines at all three gaps. Its error remains relatively small for $T=12$ and $T=16$, while the longer interval at $T=24$ produces a larger midpoint error. This behavior is expected because the endpoints provide less direct information about the field as the physical interval becomes longer. The comparison shows that the trained model can reconstruct intermediate states at temporal gaps different from a single fixed training setting, including the tested gap beyond its training range.

A related experiment on Shanghai radar precipitation is reported in Appendix~\ref{app:temporal_gap}. There, both linear interpolation and
\model{} become less accurate as the observation interval increases. At $T=4$, \model{} achieves
$21.08\%$ RFNE compared with $36.85\%$ for linear interpolation. At $T=12$, the corresponding errors are $40.42\%$ and $55.02\%$. The learned correction therefore improves the initial interpolation at both the shorter and longer tested intervals, although the remaining error increases with the temporal gap. This result is consistent with the information limitation discussed in Section~\ref{sec:problem_formulation}: widely separated observations provide less information about events occurring between them.

\subsection{Ablation Study}
\label{sec:ablation}

\begin{figure*}[t]
    \centering
    \begin{subfigure}[t]{0.32\textwidth}
        \centering
        \includegraphics[width=\linewidth]{
            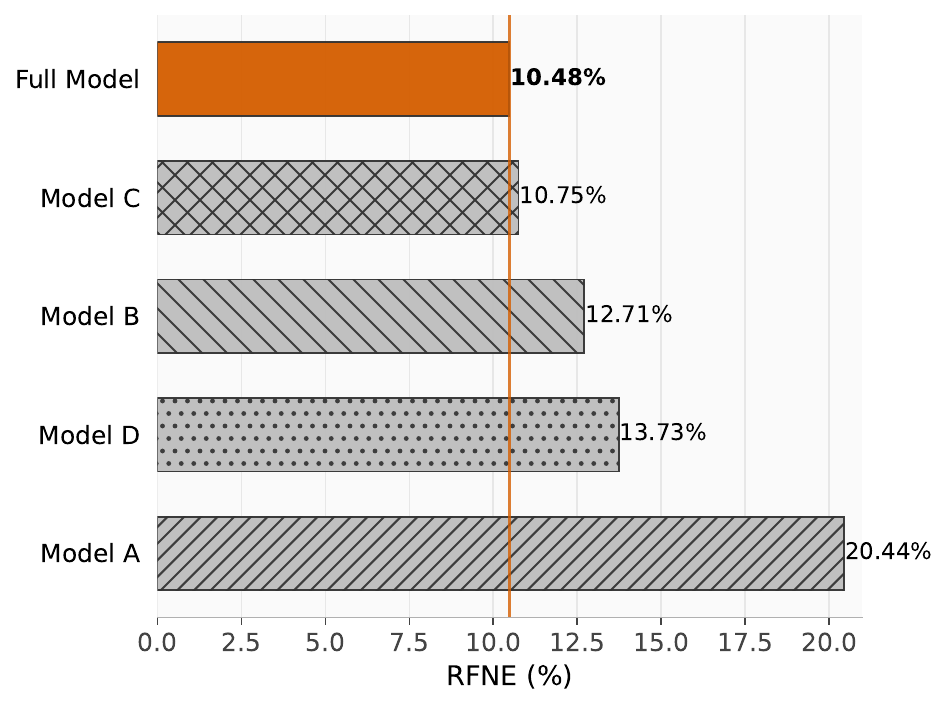
        }
        \caption{Comparison of model variants}
        \label{fig:ablation_summary}
    \end{subfigure}
    \hfill
    \begin{subfigure}[t]{0.32\textwidth}
        \centering
        \includegraphics[width=\linewidth]{
            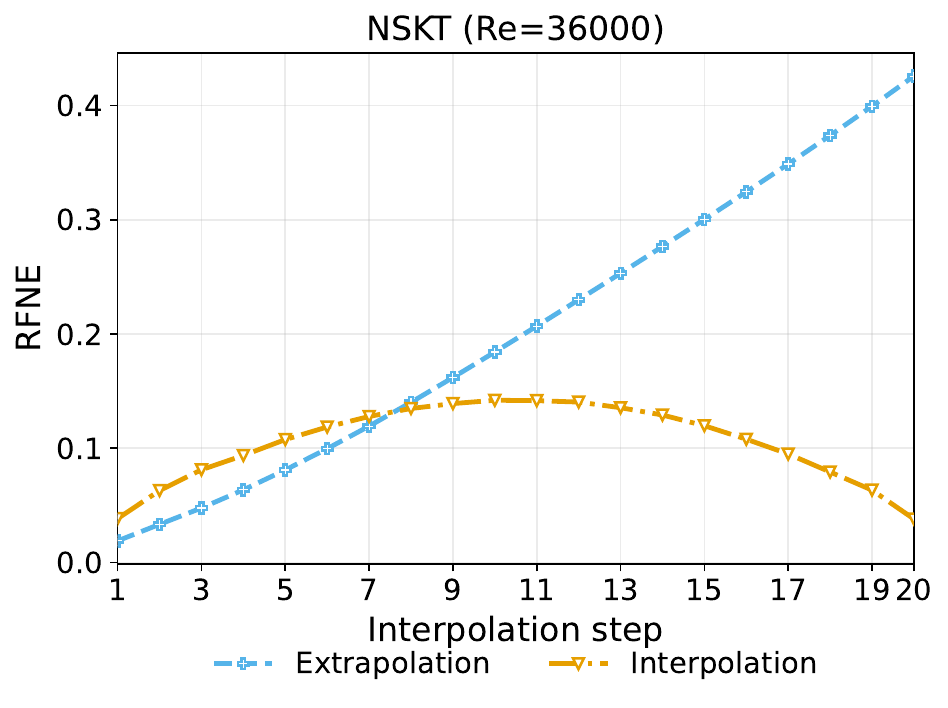
        }
        \caption{Two endpoints versus one endpoint}
        \label{fig:ablation_residual}
    \end{subfigure}
    \hfill
    \begin{subfigure}[t]{0.32\textwidth}
        \centering
        \includegraphics[width=\linewidth]{
            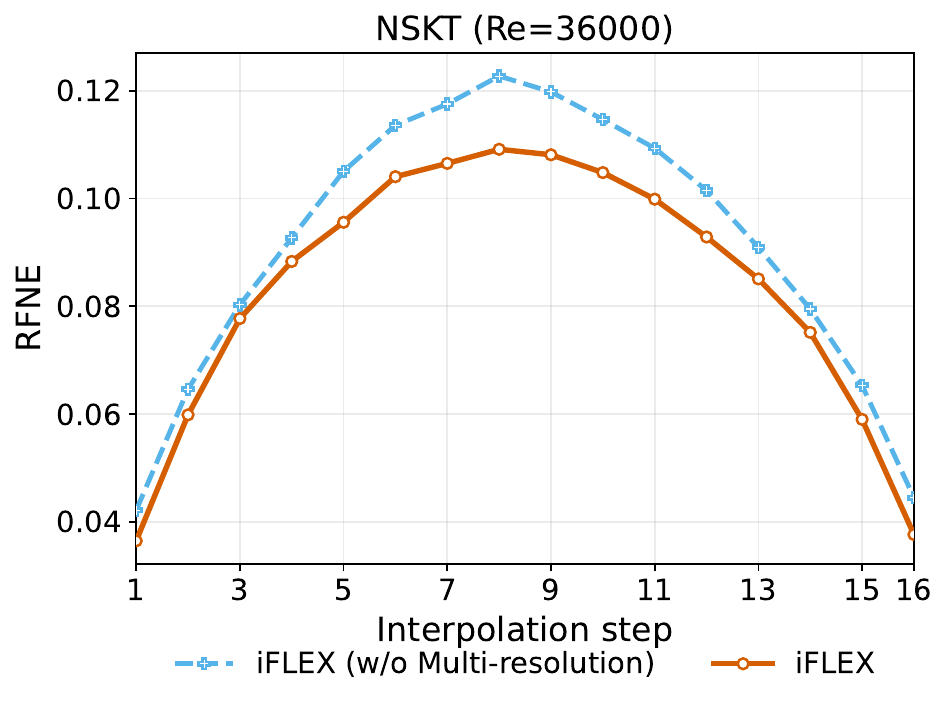
        }
        \caption{Training with different temporal gaps}
        \label{fig:ablation_multi_reso}
    \end{subfigure}
    \caption{
        Learning the interpolation residual gives the lowest reconstruction error in the NSKT ablation study at $Re=36000$. (a) The full model achieves $10.48\%$ RFNE, compared with $20.44\%$ for the one-endpoint predictor (Model A), $12.71\%$ without interval duration conditioning (Model B), $10.75\%$ with fixed-gap training (Model C), and $13.73\%$ for direct generation of the full intermediate field (Model D). (b) Model A uses only the first observation and has larger errors toward the end of the interval, whereas \model{} uses both endpoints. (c) The full model is compared with Model C, which is trained at a fixed gap of $T=20$. The curves show the error at each intermediate time.
    }
    \label{fig:ablation}
\end{figure*}

The previous experiments establish the accuracy of the complete \model{} system across different datasets and temporal conditions. We now examine which aspects of the method contribute to this performance. The central question is whether learning the residual relative to linear interpolation is more effective than generating the full intermediate field directly. We also test the role of the second endpoint, physical time conditioning, and training with different temporal gaps.
All ablations use the NSKT dataset and are evaluated at $Re=36000$. Figure~\ref{fig:ablation} summarizes the results. The full model achieves
$10.48\%$ RFNE, while each evaluated variant
has a larger reported error. The differences are not equally informative, however, because the ablations modify different parts of the reconstruction procedure.

\textbf{Learning the residual instead of the full field.} The most direct test of our interpolation formulation is Model D. This variant uses the same FLEX backbone, endpoint observations, scalar conditioning, diffusion objective, training configuration, and sampling procedure as the full model. The only difference is the quantity generated by diffusion. Model D generates the complete intermediate field $\mathbf{u}_\tau$, whereas \model{} generates
the residual $\mathbf{r}_\tau=\mathbf{u}_\tau-\widetilde{\mathbf{u}}_\tau$.
As shown in Figure~\ref{fig:ablation}(a), Model D obtains $13.73\%$ RFNE, compared with $10.48\%$ for residual generation. The corresponding SSIM values are $0.827$ and $0.869$, respectively.
Because the observations, network, and training setup remain fixed, this comparison isolates the effect of the learning target. It provides direct empirical evidence that learning the correction to a linear estimate is more accurate than generating the entire intermediate field in this setting.
This result is particularly interesting because the linear estimate is itself inaccurate at $Re=36000$, with an RFNE of $43.96\%$. The improvement therefore does not require the initial estimate to be a good standalone reconstruction. Instead, using the same estimate to define the training target and the final reconstruction gives the diffusion model a different quantity to learn. The experiment supports this formulation for the evaluated turbulence setting, although it does not establish that linear interpolation is the optimal reference for every physical system.

\textbf{Using both observed endpoints.}
Model A examines a different question. It uses only the first observation and learns the change from that observation to the target field. The later endpoint is removed from the conditioning inputs. Consequently, Model A must estimate the evolution without knowing where the system ends the interval.
Model A obtains $20.44\%$ RFNE, compared with $10.48\%$ for the complete two-endpoint model. Figure~\ref{fig:ablation}(b) shows that its error generally increases with the requested target time, while the error of \model{} is lower near both observed endpoints and largest toward the middle of the interval. This difference is consistent with the amount of information available to each model. The two-endpoint formulation can use the later observation to constrain reconstruction near the end of the interval.
This comparison changes both the available observations and the learning target. It therefore evaluates the tested one-endpoint predictor against the complete interpolation formulation. The contribution of residual learning itself is isolated by Model D.

\textbf{Conditioning on the observation interval.}
Model B removes $\Delta t$ from the scalar conditioning variables while retaining the remaining model components. Its RFNE increases to $12.71\%$, compared with $10.48\%$ for the full model. This experiment supports the distinction between the relative target position and the actual duration of the observation interval. Two reconstruction problems can have the same interpolation ratio $\tau$ but different endpoint separations, and the resulting physical evolution may differ substantially. Providing $\Delta t$ allows the network to distinguish these cases.

\textbf{Training across different temporal gaps.}
Model C tests the effect of varying the endpoint separation during training. The complete model samples $T\in\{5,\ldots,20\}$, whereas Model C uses a fixed gap of $T=20$ for every training example. The network architecture, conditioning inputs, and optimization procedure are otherwise unchanged.
The full model achieves $10.48\%$ RFNE, compared with $10.75\%$ for Model C in Figure~\ref{fig:ablation}(a). The difference is relatively small compared with the effects observed for Models A, B, and D. Figure~\ref{fig:ablation}(c) also compares reconstruction errors at individual target steps when evaluating at $T=16$. The results indicate that exposure to different temporal gaps during training can improve accuracy when requested intervals differ from fixed training gaps.

The remaining conditioning choices are examined in the appendix. In particular, Appendix~\ref{app:mean_difference_ablation} compares the full spatial feature fusion with a variant that uses only the two endpoint feature maps. The methods perform similarly at unseen Reynolds numbers within the training range, while the explicit average and difference features improve RFNE at the evaluated Reynolds numbers outside that range. This suggests that these feature combinations can help the model use endpoint information under changes in physical parameters, though the effect depends on evaluation regimes.

\section{Limitations}\label{s:limitations}

The accuracy of temporal interpolation ultimately depends on how much information the two observations provide about the intermediate evolution. Although \model{} uses both endpoints to constrain the reconstruction, some changes may remain unobserved. For example, a short rainfall event that begins and ends between two radar observations may leave little evidence in either field. A learned model may generate a plausible intermediate state, but it cannot reliably determine whether such an event actually occurred. Our approach therefore assumes that the observations are sufficiently close in time to constrain the relevant physical evolution. This assumption becomes more restrictive as the observation interval increases or the system changes more rapidly. Indeed, our temporal-gap experiments show that reconstruction errors generally increase with the separation between observations, even though \model{} continues to improve over linear interpolation in the evaluated settings. For applications that depend on short-lived events, such as flood prediction, the reconstructed fields would need to be evaluated at the temporal resolution required by the downstream task.

The choice of linear interpolation as a reference also has limitations. We use it because it is inexpensive, requires no governing-equation solver, and provides a well-defined estimate from the observed endpoints. Our experiments show that learning its residual improves reconstruction accuracy, including in the controlled comparison with direct field generation. However, we do not have a theoretical characterization of when this reference is preferable to other choices. A more accurate numerical or learned estimate might provide a better starting point, particularly over longer observation intervals, although the additional computational cost would have to be justified by improvements in the final reconstruction. Furthermore, our evaluation covers a limited number of independent realizations. The Reynolds-number sweep uses one held-out trajectory per parameter value, supplemented by a two-trajectory evaluation at three Reynolds numbers. The larger variation observed at $Re=32000$ shows that reconstruction difficulty can depend on the particular flow realization. Similarly, the held-out SST region provides evidence of geographic transfer within the evaluated dataset, but does not establish performance across arbitrary ocean regions or observation conditions.

Finally, our evaluation focuses on the accuracy of individual reconstructed fields. Although \model{} can generate multiple intermediate states within the same observation interval, each requested time is processed independently. Consequently, accurate reconstruction at individual times does not guarantee that the resulting sequence varies smoothly or reproduces temporal derivatives accurately. The model also does not explicitly enforce the governing equations or conservation laws. While the diffusion model can generate different samples conditioned on the same observations, we have not evaluated whether their variation provides calibrated uncertainty estimates. These distinctions matter when reconstructed fields are used for subsequent scientific analysis. For example, small errors in individual velocity or vorticity fields may still affect estimates of temporal derivatives or derived physical quantities. Applications that depend on such properties require additional evaluation beyond the field reconstruction metrics considered here.

\section{Conclusions}\label{s:conclusion}

We have investigated whether scientific temporal interpolation can be improved by learning to correct a simple numerical estimate of the missing field. Our results show that linear interpolation can provide a useful reference for learning, even when it is inaccurate as a standalone reconstruction. The proposed \model{} approach first interpolates the two observed fields and then uses conditional diffusion to generate the residual needed to reconstruct the intermediate state. This allows the model to correct errors in the position, shape, and values of evolving physical structures. In a controlled turbulence experiment at $Re=36000$, learning the residual reduces RFNE from $13.73\%$ for direct field generation to $10.48\%$, while improving SSIM from $0.827$ to $0.869$. Since both models use the same architecture, observations, conditioning, and training procedure, this comparison provides direct evidence for the benefit of the residual formulation in the evaluated setting.

We also evaluated \model{} on radar precipitation and sea surface temperature, where the accuracy of the initial interpolation differs considerably. For Shanghai radar, the learned correction reduces RFNE from $47.78\%$ for linear interpolation to $32.69\%$, while on the more slowly varying SST fields, the error decreases from $1.94\%$ to $1.81\%$. These results show that the same formulation can improve reconstruction both when the numerical estimate has substantial errors and when it is already relatively accurate. For scientific systems represented in the training data, \model{} can therefore be used to estimate missing fields between recorded observations, allowing a denser temporal record to be reconstructed without rerunning the underlying physical simulation.

\textbf{Future work.}
Our findings suggest two directions for further investigation.
\begin{itemize}

    \item \textbf{Choosing a better interpolation reference.}
    Linear interpolation is a convenient starting point, but it may become increasingly inaccurate as the observation interval grows. An alternative is to construct the initial estimate using a learned predictor, as in residual generative models ~\citep{yu2024diffcast}, or a numerical approximation based on the governing equations. The important question is whether a more accurate initial estimate also makes the remaining correction easier to learn. Comparing these choices across different temporal gaps would help determine when simple interpolation is sufficient and when the additional cost of a more sophisticated reference is justified.

    \item \textbf{Reconstructing consistent temporal sequences.}
    Our current model reconstructs each requested intermediate field independently. An  extension is to generate neighboring states jointly, allowing the model to account for relationships between consecutive times. This connects temporal interpolation with learning continuous evolution from discrete observations~\citep{krishnapriyan2023learning}. Such an extension would be particularly relevant when reconstructed fields are used to estimate temporal derivatives or other quantities that depend on the evolution across multiple states. Future work should study the accuracy of these derived quantities.

\end{itemize}

\subsubsection*{Broader Impact Statement}

Temporal interpolation could make sparsely stored simulation outputs more useful by providing estimates of missing intermediate states. It could also support the analysis of observational records whose temporal resolution is insufficient for particular applications, including weather and climate analysis. However, reconstructed fields cannot recover information that is entirely absent from the observations. For example, a brief rainfall event occurring between two measurements may not be reliably reconstructed, even if the generated precipitation field appears plausible. Errors of this kind could affect downstream scientific analyses or decisions. Applications with safety implications, such as flood prediction, should therefore validate reconstructed fields against available observations and assess the accuracy of relevant physical quantities under the intended operating conditions. Reconstructed fields should not be treated as equivalent to direct measurements.

\subsubsection*{Acknowledgments}
NBE would like to acknowledge support from the U.S. Department of Energy, Office of Science, Office of Advanced Scientific Computing Research, through the RAPIDS Institute under the Scientific Discovery through Advanced Computing (SciDAC) program, under Contract No. DE-AC02-05CH11231 at Lawrence Berkeley National Laboratory. The views, opinions, and/or findings expressed are those of the authors and should not be interpreted as representing the official views or policies of the U.S. Government.

\bibliographystyle{plainnat}
\bibliography{refs}

\clearpage
\appendix

\section{Implementation Details}\label{app:implementation}

This section provides the dataset specifications, evaluation metrics, and training configurations used in our experiments.
We also describe the spectral diagnostic and inference-time measurements reported in the main text.
The reconstruction method itself is presented in Section~\ref{sec:method}; here, we focus on the details needed to reproduce and interpret the results.

\subsection{Datasets}\label{app:datasets}

We evaluate \model{} on three datasets representing different types of physical evolution: simulated Navier--Stokes Kraichnan turbulence (NSKT), Shanghai radar precipitation, and sea surface temperature (SST). Table~\ref{tab:dataset_summary} summarizes their spatial resolutions, observed variables, and data splits. For each dataset, training examples are constructed from two observed fields and a known intermediate state. We use $T$ to denote the number of intermediate states between the selected endpoints at the native temporal resolution. The endpoints are therefore separated by $T+1$ native time intervals.
The values in Table~\ref{tab:dataset_summary} specify the interpolation gaps used in the main evaluation; training can involve different gaps, as described in Section~\ref{s:training_inference}. For NSKT, each sample is extracted from a $256\times256$ spatial crop of a simulation performed on a $2048\times2048$ grid. 
Reynolds number is supplied as an additional physical parameter. The Shanghai radar and SST datasets contain observed fields on $128\times128$ grids. These datasets have no corresponding Reynolds-number condition.

\begin{table*}[t]
    \centering
    \caption{
        Dataset characteristics and evaluation configurations.
        NSKT contains simulated vorticity fields, while Shanghai radar and SST contain observed precipitation and temperature fields.
        The reported $T$ values specify the number of intermediate states in the main evaluation.
        The split counts are those supplied for each dataset.
    }
    \label{tab:dataset_summary}
    \resizebox{\textwidth}{!}{
        \begin{tabular}{l c c c c c}
            \toprule
            Dataset & Evaluation $T$ & Resolution & (Training, Validation, Test) & Observed field & Evaluated $Re$ \\
            \midrule
            NSKT & 20 & $256\times256$ & $(16000,4000,2000)$ & Vorticity & $600$--$36000$ \\
            Shanghai Radar & 8 & $128\times128$ & $(1545,850,850)$ & Precipitation & -- \\
            Sea Surface Temperature & 10 & $128\times128$ & $(172765,18278,66222)$ & Temperature & -- \\
            \bottomrule
        \end{tabular}
    }
\end{table*}

\subsection{Evaluation Metrics}\label{app:eval_metrics}

We evaluate the reconstructed fields using relative Frobenius norm error (RFNE), Pearson correlation (PC), and structural similarity index (SSIM).
Each metric captures a different aspect of reconstruction accuracy.
RFNE measures errors in the field values, PC measures agreement in their spatial variations, and SSIM compares spatial structure.
We compute RFNE and PC using the predicted and ground-truth fields in their original physical units. 
For each target state, \model{} generates one stochastic reconstruction from a single Gaussian initialization. The reported RFNE, PC, and SSIM values are computed between this reconstruction and the corresponding ground-truth field and are then averaged over all target states in the evaluation set.

\paragraph{Relative Frobenius norm error.}
RFNE measures the magnitude of the reconstruction error relative to the magnitude of the observed field:
\begin{equation}
    \label{eq:rfne}
    \mathrm{RFNE}
    =
    \frac{
        \left\|\widehat{\mathbf{u}}_\tau-\mathbf{u}_\tau\right\|_F
    }{
        \left\|\mathbf{u}_\tau\right\|_F
    }.
\end{equation}
Here, $\widehat{\mathbf{u}}_\tau$ is the reconstructed field and $\mathbf{u}_\tau$ is the ground truth. An RFNE of $0.1$ indicates that the norm of the reconstruction error is one tenth of the norm of the target field. Smaller values indicate more accurate reconstruction.

\paragraph{Pearson correlation.}
PC measures whether the predicted and observed fields have similar spatial variations after subtracting their respective means:
\begin{equation}
    \label{eq:pc}
    \mathrm{PC}
    =
    \frac{
        \left\langle
            \widehat{\mathbf{u}}_\tau-\overline{\widehat{\mathbf{u}}}_\tau,\,
            \mathbf{u}_\tau-\overline{\mathbf{u}}_\tau
        \right\rangle
    }{
        \left\|
            \widehat{\mathbf{u}}_\tau-\overline{\widehat{\mathbf{u}}}_\tau
        \right\|_F
        \left\|
            \mathbf{u}_\tau-\overline{\mathbf{u}}_\tau
        \right\|_F
    }.
\end{equation}
The bars denote spatial averages over the corresponding fields.
A high PC indicates that regions of relatively large and small values occur at similar locations.
However, correlation does not measure agreement in magnitude.
A reconstruction that systematically underestimates vorticity may still have high PC, which is why we report it alongside RFNE.

\paragraph{Structural similarity.}
SSIM~\citep{wang2004image} compares the local structure of two fields using their means, variances, and covariance.
We first normalize the predicted and observed fields to a common range and compute
\begin{equation}
    \label{eq:ssim}
    \mathrm{SSIM}
    =
    \frac{
        (2\mu_{\widehat{\mathbf{u}}}\mu_{\mathbf{u}}+c_1)
        (2\sigma_{\widehat{\mathbf{u}}\mathbf{u}}+c_2)
    }{
        (\mu_{\widehat{\mathbf{u}}}^{2}+\mu_{\mathbf{u}}^{2}+c_1)
        (\sigma_{\widehat{\mathbf{u}}}^{2}+\sigma_{\mathbf{u}}^{2}+c_2)
    }.
\end{equation}
Here, $\mu_{\widehat{\mathbf{u}}}$ and $\mu_{\mathbf{u}}$ are sample means, $\sigma_{\widehat{\mathbf{u}}}^{2}$ and $\sigma_{\mathbf{u}}^{2}$ are sample variances, and $\sigma_{\widehat{\mathbf{u}}\mathbf{u}}$ is their covariance.
The constants $c_1$ and $c_2$ stabilize the calculation.
Higher SSIM indicates closer agreement in spatial structure.

\paragraph{Windowed vorticity power spectrum.}
The reconstruction metrics above summarize agreement between complete fields.
For turbulence, we also examine how vorticity fluctuations are distributed across spatial scales.
Broad structures contribute at lower wavenumbers, while smaller features contribute at higher wavenumbers.
Comparing these distributions provides additional information about which spatial scales are reproduced by the models.

The evaluated NSKT fields are $256\times256$ crops from a larger simulation.
Their opposite boundaries generally do not match, so applying a Fourier transform as though each crop were periodic would introduce artificial discontinuities.
We reduce this effect by applying a separable two-dimensional Hann window $W$, which gradually reduces the field values near the crop boundaries.
Before windowing, we subtract the window-weighted mean of the vorticity field, $\bar{\omega}_W=\sum_{m,n}W_{mn}\omega_{mn}/\sum_{m,n}W_{mn}$.
The normalized discrete Fourier transform of the resulting field is
\begin{equation}
    \label{eq:windowed_vorticity_dft}
    \widehat{\omega}_W(\boldsymbol{\kappa})
    =
    \frac{1}{256^2}
    \operatorname{DFT}
    \left[
        W\left(\omega-\bar{\omega}_W\right)
    \right](\boldsymbol{\kappa}).
\end{equation}
The indices $m,n$ refer to grid locations and $\boldsymbol{\kappa}$ denotes spatial wavenumber.
To obtain a one-dimensional spectrum, we group Fourier modes into annular bins according to their wavenumber magnitude.
The shell-integrated vorticity power is
\begin{equation}
    \label{eq:vorticity_power_spectrum}
    P_{\omega}(\kappa_j)
    =
    \frac{1}{U\,\Delta\kappa}
    \sum_{\boldsymbol{\kappa}\in\mathcal{S}_j}
    \left|\widehat{\omega}_W(\boldsymbol{\kappa})\right|^2.
\end{equation}
Here, $\mathcal{S}_j$ is the annular bin centered at $\kappa_j$ with width $\Delta\kappa$, and $U=256^{-2}\sum_{m,n}W_{mn}^{2}$ corrects for the attenuation introduced by the window.
The wavenumbers retain the physical spacing of the original simulation grid.
Since the full domain has width $2\pi$, a $256$-point crop of the $2048$-point domain has width $L_{\mathrm{crop}}=\pi/4$, corresponding to a fundamental wavenumber spacing of $\Delta\kappa=8$ in full-domain wavenumber units.
We apply the same calculation to the ground-truth and reconstructed fields.
This diagnostic measures the spatial distribution of vorticity fluctuations.
It does not measure kinetic-energy conservation or verify that the reconstructed fields satisfy the Navier--Stokes equations.

\subsection{Model and Experiment Configurations}\label{app:model_config}

All models are evaluated using the dataset splits and target fields described above.
The experiments were conducted on a workstation with four NVIDIA A100 GPUs, each with 40 GB of memory.
The baseline architectures differ in how they use the observed endpoints, so we describe their conditioning inputs and training settings separately.
These details are important when interpreting performance differences between complete models.

\paragraph{U-Net.}
The supervised U-Net~\citep{ronneberger2015u} receives the two endpoint fields as separate input channels and directly predicts the intermediate state.
It is trained with an $\ell_1$ reconstruction loss.
The architecture follows \citet{erichson2025flex}, using a base width of $128$, two residual blocks at each resolution level, and channel multipliers $(1,2,4,8)$ for both $128\times128$ and $256\times256$ inputs.
It uses dropout $0.15$, scale-shift normalization, and residual upsampling and downsampling.
Multi-head attention operates at feature resolutions $32$, $16$, and $8$.
The relative target position, endpoint separation, and available Reynolds number are encoded using Fourier features and supplied through the residual-block embedding pathway.
Thus, the model can use the requested target time and physical parameters in addition to the two spatial fields.
Table~\ref{tab:train_para_unet} reports its optimization settings.

\begin{table}[t]
    \centering
    \caption{
        Training settings for the supervised U-Net.
        All three datasets use Adam with a learning rate of $10^{-4}$; training duration and batch size vary by dataset.
    }
    \label{tab:train_para_unet}
    \resizebox{0.55\linewidth}{!}{
        \begin{tabular}{l c c c c}
            \toprule
            Dataset & Epochs & Batch size & Learning rate & Optimizer \\
            \midrule
            NSKT & 150 & 48 & $1\times10^{-4}$ & Adam \\
            Shanghai & 400 & 48 & $1\times10^{-4}$ & Adam \\
            SST & 50 & 64 & $1\times10^{-4}$ & Adam \\
            \bottomrule
        \end{tabular}
    }
\end{table}

\paragraph{Super SloMo.}
We adapt Super SloMo~\citep{jiang2018super} to single-channel scientific fields.
The method uses two U-Nets to estimate motion between the observations and reconstruct the intermediate field. The first network receives the concatenated endpoint fields and predicts bidirectional optical flows. The second network uses the endpoints, predicted flows, intermediate flow estimates, and warped fields to refine the motion estimates and predict a visibility map. The requested interpolation ratio enters through the analytical coefficients used for flow interpolation and warping. We train the model against known intermediate fields using supervised reconstruction loss and omit the VGG perceptual loss designed for natural images. The standard Super SloMo model does not receive Reynolds number. We evaluate a separately trained Reynolds-number-conditioned variant in Appendix~\ref{app:re_conditioned_baselines}. Training settings for the standard model are given in Table~\ref{tab:train_para_superslomo}.

\begin{table}[t]
    \centering
    \caption{
        Training settings for Super SloMo.
        All datasets use Adam.
        NSKT uses a learning rate of $6\times10^{-5}$, while Shanghai radar and SST use $10^{-4}$.
    }
    \label{tab:train_para_superslomo}
    \resizebox{0.55\linewidth}{!}{
        \begin{tabular}{l c c c c}
            \toprule
            Dataset & Epochs & Batch size & Learning rate & Optimizer \\
            \midrule
            NSKT & 200 & 256 & $6\times10^{-5}$ & Adam \\
            Shanghai & 400 & 128 & $1\times10^{-4}$ & Adam \\
            SST & 200 & 128 & $1\times10^{-4}$ & Adam \\
            \bottomrule
        \end{tabular}
    }
\end{table}

\paragraph{EDEN.}
EDEN~\citep{zhang2025eden} generates intermediate fields in a learned latent representation.
It first trains a conditional variational autoencoder (VAE) to encode and reconstruct fields.
A diffusion transformer (DiT)~\citep{peebles2023scalable,ma2024sit} then learns to generate target representations from the endpoint observations and requested time.
At inference, the generated latent representation is decoded into the reconstructed field.

The VAE uses $16\times16$ patches, a hidden dimension of $768$, $12$ attention heads, an MLP ratio of $4$, and a latent dimension of $16$.
We adapt the original three-channel inputs and outputs to single-channel fields.
The encoder and decoder each contain four transformer blocks with temporal attention over the start, end, and target fields.
Endpoint features serve as conditioning tokens, while the target is encoded into a low-resolution latent representation with a diagonal Gaussian posterior.

We train the VAE with an $\ell_1$ reconstruction loss and Kullback--Leibler regularization of weight $\lambda_{\mathrm{KL}}=10^{-6}$.
We omit the perceptual loss, disable reconstruction clamping, and remove image-specific loss scaling that assumes values in $[0,1]$.
These changes are necessary because normalized scientific fields need not lie within the standard image range.
We select checkpoints using validation $\ell_1$ error.

The DiT operates on the learned latent tokens with a latent dimension of $16$, transformer width $768$, $12$ transformer blocks, $12$ attention heads, and an MLP ratio of $4$.
It predicts velocity under a linear transport objective.
Conditioning includes the endpoint tokens, relative target position, and a normalized endpoint-difference statistic computed from the cosine similarity of the observations.
Embeddings for the diffusion time, target position, and endpoint difference are combined and injected through adaptive layer normalization.

At inference, the model uses two Euler steps along the linear transport path, requiring two velocity-network evaluations.
The recorded solver configuration also specifies absolute tolerance $10^{-6}$ and relative tolerance $10^{-3}$.
For the reported fixed-step Euler procedure, the step count determines the number of function evaluations.
The VAE and DiT are trained separately using the settings in Table~\ref{tab:train_para_eden}.

\begin{table}[t]
    \centering
    \caption{
        Training settings for the two stages of EDEN.
        The VAE learns the field representation, and the DiT learns to generate its latent tokens.
        Both stages use AdamW with dataset-specific schedules.
    }
    \label{tab:train_para_eden}
    \resizebox{0.70\linewidth}{!}{
        \begin{tabular}{l ccc ccc c}
            \toprule
            Dataset
            & \multicolumn{3}{c}{VAE}
            & \multicolumn{3}{c}{DiT}
            & Optimizer \\
            \cmidrule(lr){2-4}
            \cmidrule(lr){5-7}
            & Epochs & Batch & LR
            & Epochs & Batch & LR
            & \\
            \midrule
            NSKT
            & 300 & 48 & $1.25\times10^{-5}$
            & 300 & 48 & $1\times10^{-4}$
            & AdamW \\
            Shanghai
            & 300 & 256 & $5\times10^{-5}$
            & 400 & 256 & $5\times10^{-5}$
            & AdamW \\
            SST
            & 100 & 256 & $5\times10^{-5}$
            & 50 & 256 & $5\times10^{-4}$
            & AdamW \\
            \bottomrule
        \end{tabular}}
\end{table}

\paragraph{\model{}.}
The full architecture and reconstruction procedure are described in Section~\ref{sec:method}.
The denoising network uses channel widths $[64,128,256,512]$, two residual blocks at each encoder and decoder resolution, eight attention heads, and an MLP ratio of $2$.
The architecture is shared across the three datasets, with the input resolution and training configuration adjusted to the corresponding data.
We use a cosine log-signal-to-noise ratio schedule, velocity prediction, and an $\ell_1$ denoising loss. At inference, a three-step DDIM-style sampler generates the residual that is added to the linear estimate.

Training examples are selected using a maximum interpolation count $T_{\max}$ for each dataset.
The implementation samples an integer gap $T$ from the range specified by $T_{\max}/4$ and $T_{\max}$, followed by a target index $k$ drawn uniformly from $\{1,\ldots,T\}$.
The relative target position is $\tau=k/(T+1)$. This exposes the model to different endpoint separations while preserving the native spatial resolution.
Scalar conditions include the relative target position, endpoint separation, available physical parameters, and diffusion time. Spatial conditioning uses the shared endpoint encoder and the feature-fusion and gated injection modules described in Section~\ref{s:structured_condition}.
The training settings in Table~\ref{tab:train_para_flex} also apply to the ablation models unless stated otherwise.

\begin{table}[t]
    \centering
    \caption{
        Training settings for \model{}.
        All three datasets use Adam with a learning rate of $10^{-4}$.
        The denoising architecture is shared, while training duration and batch size vary by dataset.
    }
    \label{tab:train_para_flex}
    \resizebox{0.5\linewidth}{!}{
        \begin{tabular}{l c c c c}
            \toprule
            Dataset & Epochs & Batch size & Learning rate & Optimizer \\
            \midrule
            NSKT & 150 & 48 & $1\times10^{-4}$ & Adam \\
            Shanghai & 300 & 48 & $1\times10^{-4}$ & Adam \\
            SST & 50 & 128 & $1\times10^{-4}$ & Adam \\
            \bottomrule
        \end{tabular}}
\end{table}

\paragraph{Inference timing.}
We report inference time per generated intermediate field.
The measurements are normalized by the batch size and number of generated target fields, so the reported values are amortized rather than necessarily representing the latency of one isolated request. For sequential reconstruction of $K$ intermediate fields, a useful estimate of total inference cost is $t_{\mathrm{sequence}}\approx Kt_{\mathrm{frame}}$, where $K$ excludes the observed endpoints. Multiple requested times can also be processed in a batch. However, each target still requires the three denoising steps of \model{}, and the amortized cost depends on the batching configuration. The number of requested fields and available hardware capacity therefore matter when applying the method to long records.

\section{Additional Reconstruction Examples}\label{app:additional_results}

The aggregate metrics in Section~\ref{s:results} describe reconstruction accuracy over the evaluation datasets. The examples below show how these differences appear in individual fields.
We include turbulent vorticity, radar precipitation, and sea surface temperature to illustrate the different structures encountered by the interpolation methods. Each displayed target time is reconstructed independently.

\begin{figure}[!htbp]
    \centering
    \includegraphics[
        width=0.94\textwidth,
        height=0.62\textheight,
        keepaspectratio
    ]{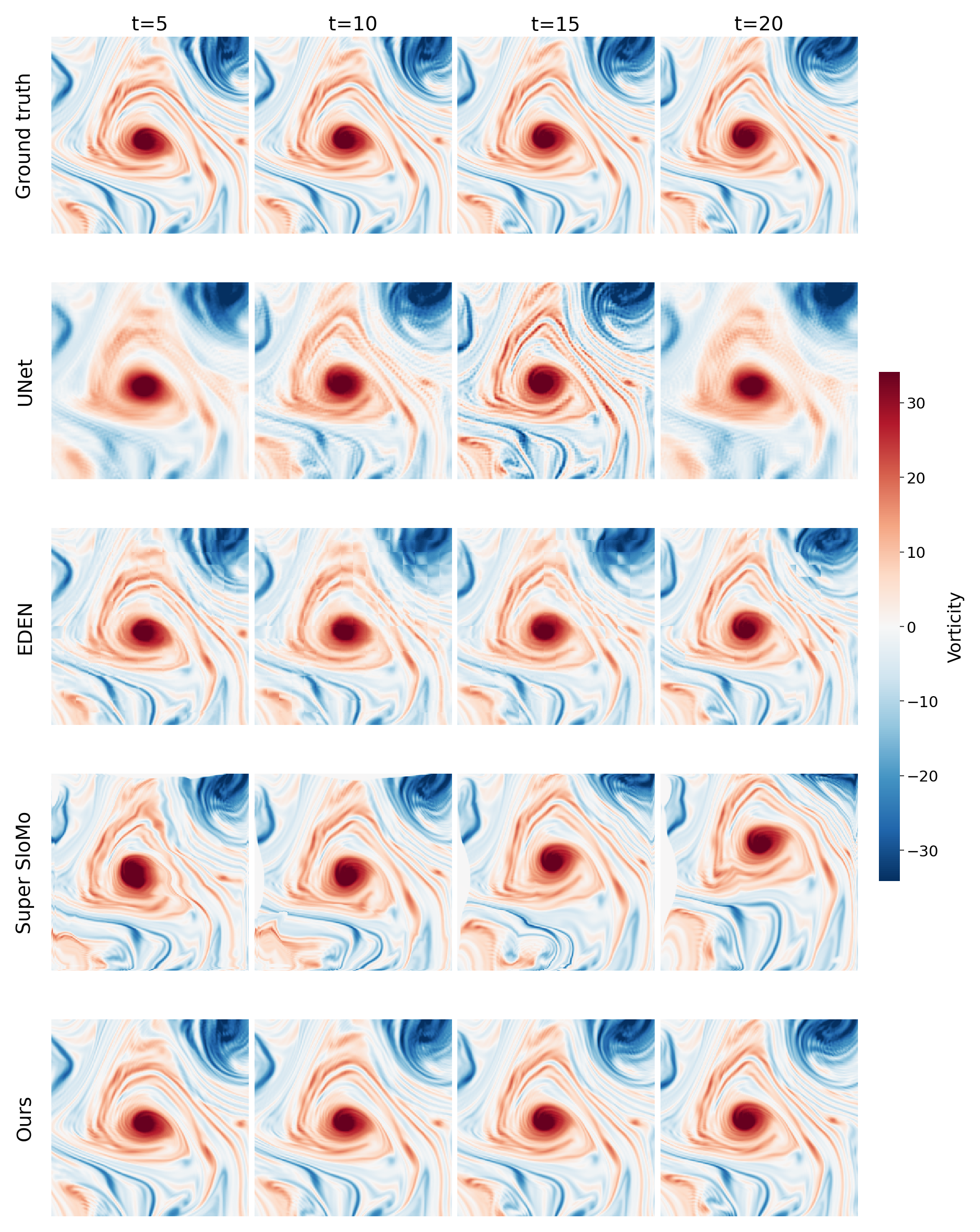}
    \caption{
        \model{} more closely reproduces the vortex structures in the illustrated NSKT fields at $Re=36000$.
        The rows show the ground truth and reconstructions from U-Net, EDEN, Super SloMo, and \model{}, evaluated at intermediate indices $k=5,10,15,20$.
        The panel labels $t$ indicate target indices, rather than physical time in seconds.
        Differences are visible in the location and shape of vortices and in smaller vorticity structures.
        Over the corresponding test set, \model{} achieves $10.48\%$ RFNE compared with $22.87\%$ for Super SloMo (Table~\ref{tab:ns_re_eval}); these aggregate values are not errors for the individual displayed fields.
    }
    \label{fig:nskt_re36000_snapshot_comparison}
\end{figure}

\begin{figure}[!htbp]
    \centering
    \includegraphics[
        width=0.94\textwidth,
        height=0.70\textheight,
        keepaspectratio
    ]{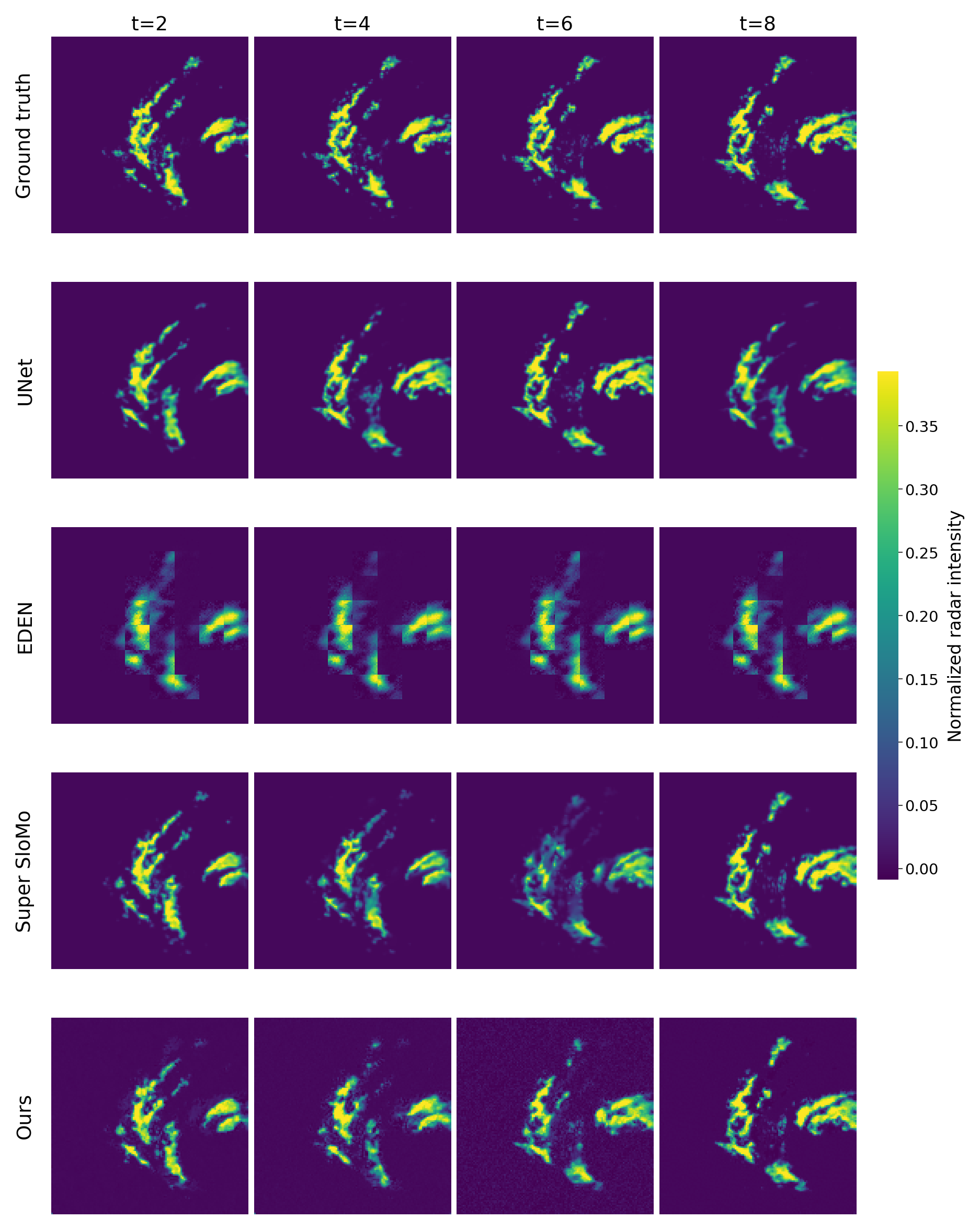}
    \caption{
        The illustrated radar reconstructions show differences in the location and intensity of localized precipitation.
        The rows compare ground-truth fields with reconstructions from U-Net, EDEN, Super SloMo, and \model{} at intermediate indices $k=2,4,6,8$.
        The panel labels $t$ denote these indices.
        In this example, \model{} more closely retains localized precipitation structures at the middle target times than EDEN or Super SloMo.
        Aggregate RFNE is $32.69\%$ for \model{}, $56.54\%$ for EDEN, and $36.77\%$ for Super SloMo (Table~\ref{tab:sh_sst_eval}).
    }
    \label{fig:shanghai_snapshot_comparison}
\end{figure}

\begin{figure}[!htbp]
    \centering
    \includegraphics[
        width=0.94\textwidth,
        height=0.70\textheight,
        keepaspectratio
    ]{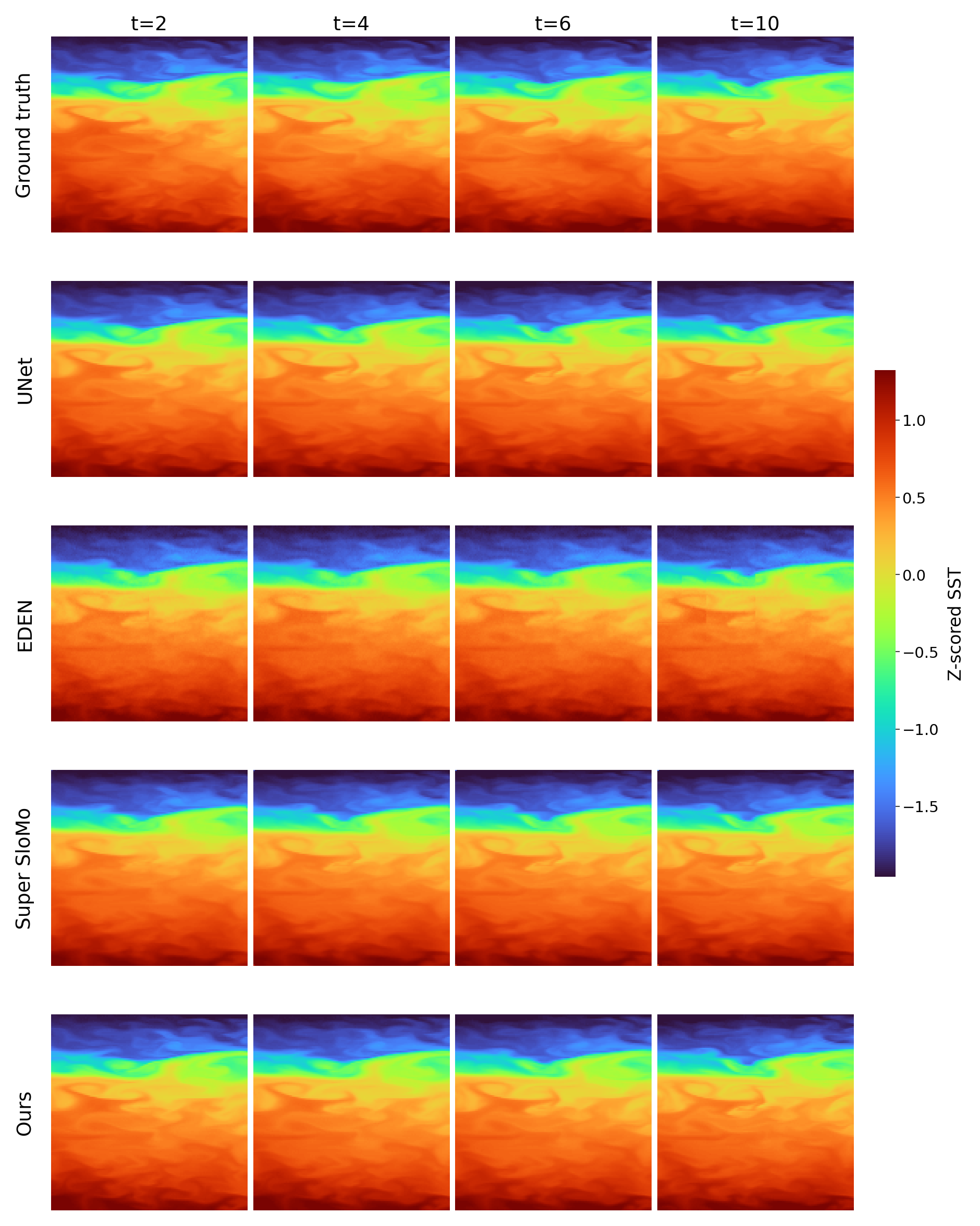}
    \caption{
        The SST example shows similar broad temperature patterns across the methods, with differences in local gradients and finer spatial variations.
        The rows compare the ground truth with U-Net, EDEN, Super SloMo, and \model{} at intermediate indices $k=2,4,6,10$.
        The panel labels $t$ denote target indices.
        On the held-out South Pacific test set, \model{} achieves $1.81\%$ RFNE and $0.87$ SSIM (Table~\ref{tab:sh_sst_eval}).
        These scores summarize the full evaluation set rather than the displayed examples.
    }
    \label{fig:sst_snapshot_comparison}
\end{figure}

\section{Refining Linear Interpolation Across Temporal Gaps}\label{app:temporal_gap}

The difficulty of temporal interpolation depends on the separation between the endpoint observations. This is particularly important for precipitation, where rain cells can move, intensify, or dissipate over relatively short periods. We use the Shanghai radar dataset to examine how the accuracy of linear interpolation and its learned correction changes with the observation interval. The temporal-gap study evaluates $T\in\{4,8,12\}$ intermediate states.
The experiment describes training at $T=8$; whether this denotes a fixed training gap or the maximum gap in variable-gap training requires confirmation. For each evaluated interval, the methods are compared at the corresponding normalized intermediate positions. Increasing $T$ increases the physical time separating the observations. For example, the midpoint estimate always gives equal weight to the endpoint fields, but over a longer interval those fields may differ much more substantially.

Table~\ref{tab:shanghai_temporal_gap} shows that reconstruction error increases with the temporal gap for both methods. At $T=4$, linear interpolation has an RFNE of $36.85\%$, which \model{} reduces to $21.08\%$. At $T=12$, the corresponding errors are $55.02\%$ and $40.42\%$.
The learned correction therefore improves the initial estimate across all three tested gaps, although the remaining error becomes larger as the interval increases.
This trend is consistent with the information limitation discussed in Section~\ref{sec:problem_formulation}. When observations are farther apart, more of the intervening evolution must be inferred. A model can learn patterns of precipitation change, but events that leave little evidence at either endpoint remain difficult to reconstruct reliably.

\begin{table}[!h]
    \centering
    \caption{
        \model{} reduces reconstruction error relative to linear interpolation at all three tested radar observation gaps.
        Both methods become less accurate as the gap increases.
        At $T=12$, residual refinement reduces RFNE from $55.02\%$ to $40.42\%$.
        Bold indicates the better result within each gap.
    }
    \label{tab:shanghai_temporal_gap}
    \resizebox{0.65\linewidth}{!}{
        \begin{tabular}{c l c c c}
            \toprule
            Intermediate states $T$ & Model & RFNE $\downarrow$ & SSIM $\uparrow$ & PC $\uparrow$ \\
            \midrule
            \multirow{2}{*}{4}
            & Linear interpolation & 36.85\% & 0.85 & 0.91 \\
            & \model{} & \textbf{21.08\%} & \textbf{0.93} & \textbf{0.97} \\
            \midrule
            \multirow{2}{*}{8}
            & Linear interpolation & 47.78\% & 0.79 & 0.84 \\
            & \model{} & \textbf{33.52\%} & \textbf{0.87} & \textbf{0.92} \\
            \midrule
            \multirow{2}{*}{12}
            & Linear interpolation & 55.02\% & 0.76 & 0.79 \\
            & \model{} & \textbf{40.42\%} & \textbf{0.84} & \textbf{0.89} \\
            \bottomrule
        \end{tabular}}
\end{table}

\section{Comparison with Supervised Temporal Interpolation}\label{app:pit_unet}

We also compare \model{} with PIT-UNet~\citep{thomas2026physics}, a supervised architecture developed for temporal interpolation.
This extends the main comparison beyond the standard U-Net and the evaluated video frame interpolation methods. PIT-UNet directly predicts an intermediate field from the two endpoint observations and the normalized target time.

An official implementation was not available for these experiments.
We therefore implement the published architecture with $64$ base channels.
The model is trained using an $\ell_1$ reconstruction loss, without the physics-based or VGG losses. We use $256\times256$ patches with stride $32$, intervals containing 20 intermediate states, and a batch size of $12$.
Optimization uses Adam for $50$ epochs with an initial learning rate of $10^{-4}$. The checkpoint with the lowest validation $\ell_1$ loss is selected for evaluation.

Table~\ref{tab:pit_unet} shows that \model{} achieves lower RFNE and higher SSIM at all three unseen Reynolds numbers.
At $Re=36000$, PIT-UNet obtains $15.94\%$ RFNE, compared with $10.48\%$ for \model{}. The corresponding SSIM values are $0.74$ and $0.87$, respectively.
These results show that the complete \model{} system improves over the implemented supervised interpolation baseline. The models differ in architecture and training objective, however, so this comparison does not isolate the effect of residual learning. That question is addressed by Model D in Section~\ref{sec:ablation}, where the learning target changes while the remaining configuration is held fixed.

\begin{table}[!h]
    \centering
    \caption{
        \model{} achieves lower RFNE and higher SSIM than the implemented PIT-UNet at three unseen Reynolds numbers.
        At $Re=36000$, RFNE decreases from $15.94\%$ to $10.48\%$.
        Both models are evaluated with 20 intermediate states per interval.
        Bold denotes the best value at the reported precision.
    }
    \label{tab:pit_unet}
    \resizebox{0.45\linewidth}{!}{
        \begin{tabular}{c l c c c}
            \toprule
            $Re$ & Model & RFNE (\%) $\downarrow$ & SSIM $\uparrow$ & PC $\uparrow$ \\
            \midrule
            \multirow{2}{*}{12000}
            & PIT-UNet & 7.81 & 0.88 & \textbf{1.00} \\
            & \model{} & \textbf{4.11} & \textbf{0.96} & \textbf{1.00} \\
            \midrule
            \multirow{2}{*}{24000}
            & PIT-UNet & 12.23 & 0.80 & 0.99 \\
            & \model{} & \textbf{7.10} & \textbf{0.92} & \textbf{1.00} \\
            \midrule
            \multirow{2}{*}{36000}
            & PIT-UNet & 15.94 & 0.74 & 0.98 \\
            & \model{} & \textbf{10.48} & \textbf{0.87} & \textbf{0.99} \\
            \bottomrule
        \end{tabular}}
\end{table}

\section{Ablation of Mean and Difference Features}\label{app:mean_difference_ablation}

The spatial conditioning module combines features from both endpoint observations with their average and signed difference. As discussed in Section~\ref{s:structured_condition}, these additional combinations contain no information beyond the original endpoint features. Their purpose is to make useful relationships between the observations directly available to the fusion module.

We examine this design by comparing the complete feature representation with an endpoint-only variant. The endpoint-only model supplies $\mathbf{f}_0^{(\ell)}$ and $\mathbf{f}_1^{(\ell)}$ to the learned fusion projection, without their explicit average and difference. The complete model uses the representation defined in Eq.~\eqref{eq:spatial_condition}. Both variants use the same observed fields, linear reference, diffusion objective, network depth and width, scalar conditions, training data, optimization, and sampling procedure. The comparison therefore examines how the endpoint features are presented to the fusion module.
The results in Table~\ref{tab:ns_re_endpoint_only_ablation} show that the effect depends on the evaluated Reynolds number. At $Re=12000$ and $24000$, which are unseen values within the training range, the endpoint-only variant achieves slightly lower RFNE. At the three evaluated Reynolds numbers outside the training range, the complete feature representation performs better.
For example, at $Re=36000$, its RFNE is $10.48\%$, compared with $11.26\%$ for endpoint-only fusion.

These results indicate that explicit mean and difference features can improve reconstruction in the tested extrapolation settings, even though they do not add observations. Their advantage is not uniform across Reynolds numbers, so the ablation supports a more limited conclusion: the representation of endpoint information can affect how the model generalizes to different physical parameters.

\begin{table*}[!h]
    \centering
    \caption{
        Explicit mean and difference features improve reconstruction at the three evaluated Reynolds numbers outside the training range, while endpoint-only fusion performs comparably at the two unseen values within it.
        Both models are trained at $Re\in\{2000,4000,8000,16000,32000\}$ and evaluated with $T=20$.
        At $Re=36000$, complete fusion achieves $10.48\%$ RFNE, compared with $11.26\%$ for endpoint-only fusion.
        Bold indicates the best value at the reported precision.
    }
    \label{tab:ns_re_endpoint_only_ablation}
    \resizebox{0.65\linewidth}{!}{
        \begin{tabular}{c l c c c}
            \toprule
            Reynolds number & Method & RFNE (\%) $\downarrow$ & PC $\uparrow$ & SSIM $\uparrow$ \\
            \midrule
            \multirow{2}{*}{$600$}
            & \model{} & \textbf{0.50} & \textbf{1.00} & \textbf{1.00} \\
            & Endpoint-only & 0.55 & 1.00 & 0.99 \\
            \midrule
            \multirow{2}{*}{$1000$}
            & \model{} & \textbf{0.73} & \textbf{1.00} & \textbf{0.99} \\
            & Endpoint-only & 0.79 & 1.00 & 0.99 \\
            \midrule
            \multirow{2}{*}{$12000$}
            & \model{} & 4.11 & 1.00 & 0.96 \\
            & Endpoint-only & \textbf{3.99} & \textbf{1.00} & \textbf{0.96} \\
            \midrule
            \multirow{2}{*}{$24000$}
            & \model{} & 7.10 & 1.00 & 0.92 \\
            & Endpoint-only & \textbf{6.92} & \textbf{1.00} & \textbf{0.93} \\
            \midrule
            \multirow{2}{*}{$36000$}
            & \model{} & \textbf{10.48} & \textbf{0.99} & \textbf{0.87} \\
            & Endpoint-only & 11.26 & 0.99 & 0.86 \\
            \bottomrule
        \end{tabular}}
\end{table*}

\section{Interpretation of Conditional Sampling}\label{app:uncertainty}

Although \model{} uses a stochastic diffusion model, variation between generated samples should not be interpreted automatically as calibrated physical uncertainty.
The model conditions on the available observations, which may not uniquely determine the intermediate state.
For example, an NSKT crop does not contain the complete flow field, and unobserved flow outside the crop may influence its evolution.
Similarly, observational records may omit relevant changes or contain measurement noise.
Different samples can therefore represent different possible reconstructions given incomplete information.

Our experiments evaluate agreement with known intermediate fields.
They do not establish that the variation between generated samples correctly represents the frequency or magnitude of reconstruction errors.
Nor do they separate uncertainty due to incomplete observations from uncertainty in the learned model.
Because each target time is generated independently, the sampling procedure also does not enforce a coherent joint evolution across consecutive reconstructed states.
Evaluating uncertainty calibration and temporal coherence would require additional experiments beyond the field reconstruction metrics reported here.

\section{Complete NSKT Evaluation Across Reynolds Numbers}\label{app:nskt_complete_re}

The main experiments report results at three unseen Reynolds numbers.
Here, we provide the complete NSKT evaluation across the parameter values used in training and those reserved for testing.
All learned models are trained at $Re\in\{2000,4000,8000,16000,32000\}$ using simulation seed $2150$. Evaluation uses a held-out trajectory generated with seed $3407$ at each Reynolds number. Each trajectory contributes 256 spatiotemporal samples, evaluated with $T=20$ intermediate states.
All methods use the same target fields, spatial sampling, preprocessing, and evaluation metrics.

Testing at a Reynolds number included in training measures reconstruction on a new realization under a familiar physical parameter. Testing at unseen values within the training range examines interpolation in parameter space, while values below or above that range test extrapolation.
These categories should not be confused with temporal interpolation: the task remains reconstruction between two observed states in every case.
Tables~\ref{tab:nskt_seen_re}--\ref{tab:nskt_extrapolation_re} report the complete results. \model{} achieves the lowest RFNE and highest SSIM across all evaluated Reynolds numbers. The reconstruction difficulty is not determined by the distance from the training parameter range alone. For example, the unseen low-Reynolds-number cases have small errors because their observed fields are comparatively smooth, whereas the upper-range case at $Re=36000$ is substantially more difficult.
The evaluation uses one held-out trajectory per Reynolds number, so these results do not describe the full variation across initial conditions.
Appendix~\ref{sec:trajectory_replication} examines additional realizations at three selected values.

\begin{table*}[!h]
    \centering
    \caption{
        \model{} achieves the lowest RFNE and highest SSIM at every Reynolds number used during training, when evaluated on independent held-out trajectories.
        RFNE ranges from $1.21\%$ at $Re=2000$ to $9.35\%$ at $Re=32000$.
        All methods use $256\times256$ crops and $T=20$.
        Bold denotes the best value at the reported precision.
    }
    \label{tab:nskt_seen_re}
    \small
    \setlength{\tabcolsep}{8pt}
    \renewcommand{\arraystretch}{0.94}
    \begin{tabular}{c l c c c}
        \toprule
        \(Re\) & Method & RFNE (\%) \(\downarrow\) & SSIM \(\uparrow\) & PC \(\uparrow\) \\
        \midrule
        \multirow{6}{*}{2000}
        & Linear interpolation & 10.97 & 0.805 & 0.991 \\
        & U-Net               & 20.85 & 0.725 & 0.983 \\
        & EDEN                & 3.11  & 0.948 & 0.999 \\
        & Super SloMo         & 4.36  & 0.956 & 0.999 \\
        & PIT-UNet            & 2.76  & 0.964 & \textbf{1.000} \\
        & \model{}            & \textbf{1.21} & \textbf{0.989} & \textbf{1.000} \\
        \midrule
        \multirow{6}{*}{4000}
        & Linear interpolation & 18.44 & 0.651 & 0.976 \\
        & U-Net               & 23.87 & 0.624 & 0.968 \\
        & EDEN                & 5.05  & 0.920 & 0.998 \\
        & Super SloMo         & 6.80  & 0.922 & 0.997 \\
        & PIT-UNet            & 3.81  & 0.946 & 0.999 \\
        & \model{}            & \textbf{1.88} & \textbf{0.981} & \textbf{1.000} \\
        \midrule
        \multirow{6}{*}{8000}
        & Linear interpolation & 26.52 & 0.506 & 0.952 \\
        & U-Net               & 29.60 & 0.515 & 0.948 \\
        & EDEN                & 8.84  & 0.866 & 0.995 \\
        & Super SloMo         & 10.25 & 0.867 & 0.994 \\
        & PIT-UNet            & 5.85  & 0.912 & 0.998 \\
        & \model{}            & \textbf{2.77} & \textbf{0.973} & \textbf{0.999} \\
        \midrule
        \multirow{6}{*}{16000}
        & Linear interpolation & 34.57 & 0.390 & 0.920 \\
        & U-Net               & 36.49 & 0.415 & 0.920 \\
        & EDEN                & 15.74 & 0.771 & 0.985 \\
        & Super SloMo         & 15.13 & 0.790 & 0.986 \\
        & PIT-UNet            & 9.44  & 0.847 & 0.994 \\
        & \model{}            & \textbf{4.23} & \textbf{0.958} & \textbf{0.999} \\
        \midrule
        \multirow{6}{*}{32000}
        & Linear interpolation & 42.55 & 0.311 & 0.884 \\
        & U-Net               & 44.41 & 0.358 & 0.887 \\
        & EDEN                & 24.15 & 0.645 & 0.966 \\
        & Super SloMo         & 21.51 & 0.688 & 0.973 \\
        & PIT-UNet            & 14.86 & 0.753 & 0.987 \\
        & \model{}            & \textbf{9.35} & \textbf{0.885} & \textbf{0.994} \\
        \bottomrule
    \end{tabular}
\end{table*}

\begin{table*}[!h]
    \centering
    \caption{
        \model{} achieves the lowest reconstruction error at both unseen Reynolds numbers within the training range.
        At $Re=12000$ and $24000$, its RFNE is $4.11\%$ and $7.10\%$, compared with $7.81\%$ and $12.23\%$ for PIT-UNet.
        All methods use $256\times256$ crops and $T=20$.
        Bold denotes the best value at the reported precision.
    }
    \label{tab:nskt_interior_re}
    \small
    \setlength{\tabcolsep}{8pt}
    \renewcommand{\arraystretch}{0.96}
    \begin{tabular}{c l c c c}
        \toprule
        \(Re\) & Method & RFNE (\%) \(\downarrow\) & SSIM \(\uparrow\) & PC \(\uparrow\) \\
        \midrule
        \multirow{6}{*}{12,000}
        & Linear interpolation & 32.12 & 0.433 & 0.932 \\
        & U-Net               & 39.54 & 0.414 & 0.921 \\
        & EDEN                & 13.15 & 0.815 & 0.989 \\
        & Super SloMo         & 13.20 & 0.826 & 0.989 \\
        & PIT-UNet            & 7.81  & 0.880 & 0.996 \\
        & \model{}            & \textbf{4.11} & \textbf{0.957} & \textbf{0.999} \\
        \midrule
        \multirow{6}{*}{24,000}
        & Linear interpolation & 39.55 & 0.346 & 0.898 \\
        & U-Net               & 50.97 & 0.269 & 0.876 \\
        & EDEN                & 20.14 & 0.708 & 0.976 \\
        & Super SloMo         & 18.38 & 0.741 & 0.980 \\
        & PIT-UNet            & 12.23 & 0.801 & 0.991 \\
        & \model{}            & \textbf{7.10} & \textbf{0.919} & \textbf{0.997} \\
        \bottomrule
    \end{tabular}
\end{table*}

\begin{table*}[!h]
    \centering
    \caption{
        \model{} achieves the lowest RFNE for all three evaluated Reynolds numbers outside the training range.
        Its errors are $0.50\%$ at $Re=600$, $0.73\%$ at $Re=1000$, and $10.48\%$ at $Re=36000$.
        The larger error in the upper-range case reflects the greater reconstruction difficulty of the tested flow, rather than parameter extrapolation distance alone.
        All methods use $256\times256$ crops and $T=20$.
    }
    \label{tab:nskt_extrapolation_re}
    \small
    \setlength{\tabcolsep}{8pt}
    \renewcommand{\arraystretch}{0.96}
    \begin{tabular}{c c l c c c}
        \toprule
        \(Re\) & Direction & Method & RFNE (\%) \(\downarrow\) & SSIM \(\uparrow\) & PC \(\uparrow\) \\
        \midrule
        \multirow{6}{*}{600}
        & \multirow{6}{*}{Lower}
        & Linear interpolation & 2.18  & 0.977 & \textbf{1.000} \\
        && U-Net               & 43.19 & 0.622 & 0.990 \\
        && EDEN                & 1.55  & 0.970 & \textbf{1.000} \\
        && Super SloMo         & 1.80  & 0.986 & \textbf{1.000} \\
        && PIT-UNet            & 2.35  & 0.972 & \textbf{1.000} \\
        && \model{}            & \textbf{0.50} & \textbf{0.995} & \textbf{1.000} \\
        \midrule
        \multirow{6}{*}{1,000}
        & \multirow{6}{*}{Lower}
        & Linear interpolation & 4.55  & 0.932 & 0.998 \\
        && U-Net               & 21.54 & 0.810 & 0.994 \\
        && EDEN                & 1.81  & 0.966 & \textbf{1.000} \\
        && Super SloMo         & 2.54  & 0.978 & \textbf{1.000} \\
        && PIT-UNet            & 2.26  & 0.971 & \textbf{1.000} \\
        && \model{}            & \textbf{0.73} & \textbf{0.993} & \textbf{1.000} \\
        \midrule
        \multirow{6}{*}{36,000}
        & \multirow{6}{*}{Upper}
        & Linear interpolation & 43.96 & 0.301 & 0.876 \\
        && U-Net               & 59.94 & 0.181 & 0.833 \\
        && EDEN                & 25.52 & 0.626 & 0.962 \\
        && Super SloMo         & 22.87 & 0.673 & 0.969 \\
        && PIT-UNet            & 15.94 & 0.737 & 0.985 \\
        && \model{}            & \textbf{10.48} & \textbf{0.869} & \textbf{0.993} \\
        \bottomrule
    \end{tabular}
\end{table*}

\section{Reynolds-Number Conditioning for Baseline Models}\label{app:re_conditioned_baselines}

The standard Super SloMo and EDEN baselines do not receive Reynolds number in the same way as \model{}. This creates a difference in the information available to the models when reconstructing NSKT fields. To examine its effect, we train additional versions of EDEN and Super SloMo that receive Reynolds number as a conditioning input.
Both variants are trained using $Re\in\{2000,4000,8000,16000,32000\}$ and evaluated at the unseen values in Table~\ref{tab:re_conditioned_baselines}.
The experiments use $256\times256$ crops and $T=20$.
Reynolds number is normalized as $r_{\mathrm{Re}}=Re/40000$, but the conditioning mechanism differs between the two architectures.
EDEN already has a global conditioning pathway, so we apply a learned linear embedding to $r_{\mathrm{Re}}$ and add it to the embeddings of diffusion time, target position, and endpoint difference. Super SloMo does not have an equivalent scalar-conditioning pathway. We therefore broadcast $r_{\mathrm{Re}}$ over the spatial grid and concatenate it as a constant input channel to both flow-estimation and flow-refinement networks. The remaining architecture of each baseline is unchanged.

Table~\ref{tab:re_conditioned_baselines} shows that the additional condition affects the two methods differently.
For EDEN, Reynolds-number conditioning reduces RFNE by approximately $7.2\%$--$19.4\%$ relative, depending on the test case.
For Super SloMo, the constant-channel condition increases RFNE by approximately $6.1\%$--$8.3\%$ under the tested configuration.
This does not imply that Reynolds number is unhelpful to motion-based interpolation in general; the result depends on how the information enters the model and how that variant is trained.
\model{} retains the lowest RFNE at every evaluated Reynolds number.
At $Re=36000$, its error is $10.48\%$, compared with $23.67\%$ for conditioned EDEN and $24.26\%$ for conditioned Super SloMo.
These additional comparisons show that the performance difference in the main results cannot be explained simply by the omission of Reynolds number from the standard baselines.

\begin{table*}[!h]
    \centering
    \caption{
        \model{} retains the lowest RFNE when EDEN and Super SloMo are additionally conditioned on Reynolds number.
        Conditioning improves EDEN under the tested implementation, while the constant-channel condition increases Super SloMo's error.
        At $Re=36000$, the conditioned baselines achieve $23.67\%$ and $24.26\%$ RFNE, respectively, compared with $10.48\%$ for \model{}.
        Bold indicates the lowest error at each Reynolds number.
    }
    \label{tab:re_conditioned_baselines}
    \resizebox{0.7\linewidth}{!}{
        \begin{tabular}{c c c c c c}
            \toprule
            \(Re\) & \model{} & EDEN & EDEN+\(Re\) & Super SloMo & Super SloMo+\(Re\) \\
            \midrule
            600
            & \textbf{0.50\%} & 1.55\% & 1.27\% & 1.80\% & 1.95\% \\
            1,000
            & \textbf{0.73\%} & 1.81\% & 1.46\% & 2.54\% & 2.70\% \\
            12,000
            & \textbf{4.11\%} & 13.15\% & 11.67\% & 13.20\% & 14.22\% \\
            24,000
            & \textbf{7.10\%} & 20.14\% & 18.43\% & 18.38\% & 19.85\% \\
            36,000
            & \textbf{10.48\%} & 25.52\% & 23.67\% & 22.87\% & 24.26\% \\
            \bottomrule
        \end{tabular}
    }
\end{table*}

\section{Evaluation Across Additional Test Realizations}\label{sec:trajectory_replication}

The complete Reynolds-number evaluation uses one held-out NSKT trajectory at each parameter value.
Although each trajectory provides many spatiotemporal samples, those samples share the same simulated realization.
They therefore do not measure how reconstruction accuracy varies between different initial conditions at a fixed Reynolds number.
To examine this variation, we evaluate the same trained checkpoint on a second held-out trajectory at $Re=8000$, $12000$, and $32000$.
The values $Re=8000$ and $32000$ are included in training, while $Re=12000$ is unseen but lies within the training parameter range.
The original test seed is $3407$. We add seed $1000$ at $Re=8000$ and $12000$, and seed $987$ at $Re=32000$. All test seeds differ from the training seed $2150$.
Each trajectory contributes 256 spatiotemporal samples, evaluated with 20 intermediate target states between the observations.
We first average RFNE over samples and target times within each trajectory.
Table~\ref{tab:nskt_trajectory_robustness} then reports the mean and sample standard deviation of the two trajectory-level errors at each Reynolds number.
Thus, the reported standard deviation describes variation between realizations, rather than variation between spatial crops of one realization.
The reported variation is smaller at $Re=8000$ and $12000$ than at $Re=32000$.
At the highest evaluated value, the two realizations appear to differ substantially in reconstruction difficulty.
This finding is relevant to the interpretation of the Reynolds-number sweep: performance at a particular parameter value also depends on the flow realization.
With only two trajectories per value, the experiment provides an initial check rather than a precise estimate of variation across initial conditions.

\begin{table}[!h]
    \centering
    \caption{
        Reconstruction error varies more between the two evaluated trajectories at $Re=32000$ than at $Re=8000$ or $12000$.
        Values report the mean and sample standard deviation of trajectory-level RFNE for a fixed trained checkpoint.
        Each trajectory contributes 256 spatiotemporal samples with 20 intermediate targets.
        The two realizations per Reynolds number provide a limited estimate of sensitivity to initial conditions.
    }
    \label{tab:nskt_trajectory_robustness}
    \resizebox{0.35\linewidth}{!}{
    \begin{tabular}{c c c}
        \toprule
        $Re$ & Test seeds & RFNE (\%) $\downarrow$ \\
        \midrule
        8,000  & 3407, 1000 & $2.93 \pm 0.35$ \\
        12,000 & 3407, 1000 & $4.29 \pm 0.74$ \\
        32,000 & 3407, 987  & $6.14 \pm 3.84$ \\
        \bottomrule
    \end{tabular}}
\end{table}

\end{document}